\documentclass[10pt,twocolumn,letterpaper]{article}

\usepackage[pagenumbers]{cvpr} 

\usepackage[dvipsnames]{xcolor}

\definecolor{cvprblue}{rgb}{0.21,0.49,0.74}

\usepackage{array}
\usepackage{subcaption}
\usepackage{multirow,multicol}
\usepackage[flushleft]{threeparttable}
\usepackage{comment}
\usepackage{algorithmic}
\usepackage{booktabs}
\usepackage{bbm, dsfont}
\usepackage[font=small,labelfont=bf]{caption}

\usepackage{amssymb}
\usepackage{pifont}

\newcolumntype{x}[1]{>{\centering\arraybackslash}p{#1pt}}
\newcommand{\app}{\raise.17ex\hbox{$\scriptstyle\sim$}}

\newlength\savewidth

\makeatletter\renewcommand\paragraph{\@startsection{paragraph}{4}{\z@}
  {.5em \@plus1ex \@minus.2ex}{-.5em}{\normalfont\normalsize\bfseries}}\makeatother

\def\tablecite#1#{%
  \def\pretablecite{#1}%
  \tableciteaux}
\def\tableciteaux#1{%
  \textsuperscript{\expandafter\originalcite\pretablecite{#1}}%
}

\usepackage{graphicx}
\usepackage{enumitem}
\usepackage{wrapfig}
\usepackage{lipsum}
\usepackage{soul}

\usepackage{color, colortbl}

\usepackage{tabu}
\usepackage{xcolor}
\usepackage{nicematrix}

\definecolor{ForestGreen}{rgb}{0.13, 0.55, 0.13}
\definecolor{Green}{rgb}{0.0, 0.5, 0.0}
\definecolor{Blue}{rgb}{0.25, 0.42, 0.88}
\definecolor{green(munsell)}{rgb}{0.0, 0.66, 0.47}
\definecolor{green(ryb)}{rgb}{0.4, 0.69, 0.2}
\definecolor{green(pigment)}{rgb}{0.0, 0.65, 0.31}
\definecolor{citecolor}{HTML}{0071bc}
\definecolor{GrayXMark}{gray}{0.7}
\definecolor{DifferenceColor}{HTML}{af3235}
\definecolor{HighlightColor}{gray}{0.9}
\definecolor{OracleTextColor}{gray}{0.55}
\definecolor{Cerulean}{HTML}{00a2e3}

\usepackage{xcolor}
\definecolor{myred}{RGB}{255, 0, 0}
\definecolor{mygreen}{RGB}{0, 255, 0}
\definecolor{myblue}{RGB}{0, 0, 255}
\definecolor{myyellow}{RGB}{255, 255, 0}
\definecolor{mycyan}{RGB}{0, 255, 255}
\definecolor{mymagenta}{RGB}{255, 0, 255}
\definecolor{myorange}{RGB}{255, 165, 0}
\definecolor{mypurple}{RGB}{128, 0, 128}
\definecolor{mylime}{RGB}{0, 255, 0}
\definecolor{mypink}{RGB}{255, 192, 203}
\definecolor{mybrown}{RGB}{165, 42, 42}
\definecolor{myolive}{RGB}{128, 128, 0}
\definecolor{myteal}{RGB}{0, 128, 128}
\definecolor{mynavy}{RGB}{0, 0, 128}
\definecolor{mymaroon}{RGB}{128, 0, 0}
\definecolor{myaquamarine}{RGB}{127, 255, 212}
\definecolor{mygold}{RGB}{255, 215, 0}
\definecolor{mycrimson}{RGB}{220, 20, 60}
\definecolor{myorchid}{RGB}{218, 112, 214}
\definecolor{myindigo}{RGB}{75, 0, 130}
\definecolor{mysalmon}{RGB}{250, 128, 114}
\definecolor{mysienna}{RGB}{160, 82, 45}
\definecolor{mytan}{RGB}{210, 180, 140}
\definecolor{mylavender}{RGB}{230, 230, 250}
\definecolor{myturquoise}{RGB}{64, 224, 208}
\definecolor{mycoral}{RGB}{255, 127, 80}
\definecolor{mysilver}{RGB}{192, 192, 192}
\definecolor{myskyblue}{RGB}{135, 206, 235}
\definecolor{myplum}{RGB}{221, 160, 221}
\definecolor{mydarkgreen}{RGB}{0, 100, 0}

\definecolor{coloredspace}{RGB}{240,240,240}

\newcolumntype{H}{>{\setbox0=\hbox\bgroup}c<{\egroup}@{}}
\newcolumntype{a}{>{\columncolor{HighlightColor}}c}
\newcolumntype{L}[1]{>{\centering\arraybackslash}m{#1}}

\usepackage{tabularx}
\usepackage[export]{adjustbox}
\usepackage{makecell}
\usepackage{ragged2e}

\usepackage{dblfloatfix}

\usepackage[pagebackref,breaklinks,colorlinks,citecolor=cvprblue]{hyperref}

\usepackage[capitalize]{cleveref}
\crefname{section}{\S}{\S\S}
\crefname{subsection}{\S}{\S\S}
\crefformat{table}{Table~#2#1#3}
\crefformat{figure}{Figure~#2#1#3}
\crefformat{equation}{Eq~#2#1#3}

\title{Grounding Text-to-Image Diffusion Models for Controlled High-Quality Image Generation}

\author{
{\normalsize Ahmad Süleyman}\\
\vspace{-0.3em}
{\normalsize Department of Computer Engineering}\\
{\normalsize Turkish-German University}
\and
{\normalsize Göksel Biricik}\\
\vspace{-0.3em}
{\normalsize Department of Computer Engineering}
\\{\normalsize Yıldız Technical University}
}

\begin{document}

\twocolumn[{%
  \renewcommand\twocolumn[1][]{#1}%
  \maketitle
    \vspace{-12pt}
    \captionsetup{type=figure}
    
    \vspace{18pt}
}]

\begin{abstract}
Text-to-image (T2I) generative diffusion models have demonstrated outstanding performance in synthesizing diverse, high-quality visuals from text captions. Several layout-to-image models have been developed to control the generation process by utilizing a wide range of layouts, such as segmentation maps, edges, and human keypoints. In this work, we propose ObjectDiffusion, a model that conditions T2I diffusion models on semantic and spatial grounding information, enabling the precise rendering and placement of desired objects in specific locations defined by bounding boxes. To achieve this, we make substantial modifications to the network architecture introduced in ControlNet to integrate it with the grounding method proposed in GLIGEN. We fine-tune ObjectDiffusion on the COCO2017 training dataset and evaluate it on the COCO2017 validation dataset. Our model improves the precision and quality of controllable image generation, achieving an AP$_{\text{50}}$ of 46.6, an AR of 44.5, and an FID of 19.8, outperforming the current SOTA model trained on open-source datasets across all three metrics. ObjectDiffusion demonstrates a distinctive capability in synthesizing diverse, high-quality, high-fidelity images that seamlessly conform to the semantic and spatial control layout. Evaluated in qualitative and quantitative tests, ObjectDiffusion exhibits remarkable grounding capabilities in closed-set and open-set vocabulary settings across a wide variety of contexts. The qualitative assessment verifies the ability of ObjectDiffusion to generate multiple detailed objects in varying sizes, forms, and locations.
\end{abstract}

\section{Introduction}
Image generation is a prominent subfield within the rapidly evolving domains of Generative Artificial Intelligence (GenAI) and Deep Learning (DL). Due to their transformative practical applications, Generative Adversarial Networks (GANs) \cite{goodfellow2014generative} have attracted considerable attention in both academia and industry. Diffusion-based image generation models \cite{midjourney, dalle3, rombach2022high, saharia2022photorealistic} have emerged as powerful base models, capable of synthesizing stunningly detailed realistic images of high-quality, spanning a diverse array of concepts and domains.

One major challenge hindering the widespread adoption of generative AI, in general, and image generation models, in particular, is the difficulty of customization. Customization is an active research area focused on steering the image generation process towards producing controllable content that meets user-specified requirements. While text prompts \cite{liu2022design, pavlichenko2023best} are the most common type for controlling image generation models, they are often deemed inadequate when a high degree of control over content and style is required in the synthesized image. Specifically, text captions fall short in providing precise descriptions of the sizes, colors, shapes, textures, spatial alignments, and interactions between objects in the generated image. This limitation has prompted the adaptation and integration of more effective control formats, alongside the caption, to achieve a greater degree of controllability over the output scenes. Well-established conditioning modalities, such as segmentation maps \cite{zhou2017scene}, normal maps \cite{vasiljevic2019diode}, depth maps \cite{ranftl2020towards}, edge maps \cite{canny1986computational, xie2015holistically}, human keypoints \cite{cao2017realtime}, bounding boxes \cite{redmon2016you}, user scribbles, cartoon drawings, image styles, or combinations thereof, have been leveraged to tailor image generation. 

In this work, we present ObjectDiffusion, a conditional image generative model that augments text-to-image (T2I) models \cite{nichol2021glide, ramesh2021zero} by incorporating new control modalities. ObjectDiffusion can condition T2I models on object detection annotations \cite{redmon2016you}. Specifically, in addition to text prompts, we ground T2I models using object names and their corresponding bounding boxes. The object names are not limited to a predefined set of categories; rather, they can be in an open-ended text format, for example, \textit{a striped pink full-face motorcycle helmet}. ObjectDiffusion consists of two pretrained networks. The first is a fixed T2I Stable Diffusion (SD) \cite{rombach2022high} model, pretrained on filtered high-resolution images from the LAION-5B \cite{schuhmann2022laion} dataset. The second module is GroundNet, a parallel, trainable network that consists of the pretrained encoding blocks and the middle block of GLIGEN \cite{li2023gligen}. GroundNet is pretrained on the Object365 \cite{shao2019objects365}, GoldG (Flickr and VG) \cite{li2022grounded}, SBU \cite{ordonez2011im2text}, and CC3M \cite{sharma2018conceptual} datasets but is not fine-tuned on the COCO \cite{lin2014microsoft} dataset. GroundNet is integrated into the locked SD \cite{rombach2022high} base model via zero convolution layers.

Our approach integrates the conditional architecture design of ControlNet with the grounding method established in GLIGEN by seamlessly leveraging their design concepts and techniques to enhance grounding capabilities while improving image quality. The overall architecture of our proposed model is inspired by ControlNet, allowing optimal use of pretrained knowledge while enabling fine-tuning with new control inputs. However, since ControlNet does not support the conditional layout of entity names and bounding boxes, the architecture and initialization of GroundNet are adopted from GLIGEN to process and integrate the object names and bounding boxes information. We fine-tune ObjectDiffusion on the COCO2017 \cite{lin2014microsoft} training dataset and conduct both quantitative and qualitative evaluations on the COCO2017 \cite{lin2014microsoft} validation dataset.

The contributions of our work can be summarized as follows:

\begin{itemize} 
    \item We propose a cutting-edge model, ObjectDiffusion, and demonstrate its ability to ground pretrained T2I models using open-ended object descriptions and bounding box annotations.
    
    \item ObjectDiffusion combines the network architecture of ControlNet \cite{zhang2023adding} with the grounding technique of GLIGEN \cite{li2023gligen}, enhancing both image quality and controllability.
    
    \item ObjectDiffusion outperforms existing state-of-the-art models trained on open-source datasets, achieving superior performance on FID \cite{heusel2017gans}, AP$_{\text{50}}$ \cite{powers2020evaluation}, and AR \cite{powers2020evaluation} metrics. Our model generates high-quality, controllable images.
    
    \item We significantly reduce training costs by effectively leveraging the generation and grounding capabilities of two pretrained models: SD \cite{rombach2022high} and GLIGEN \cite{li2023gligen}.
    
    \item We evaluate the performance of our model through both quantitative and qualitative assessments, providing a comprehensive analysis of the results.

\end{itemize}

\section{Related Work}
 
\subsection{Diffusion Models} 
Sohl-Dickstein et al. \cite{sohl2015deep} established the theoretical foundation of probabilistic diffusion models, a framework in which random noise is iteratively added to ground-truth images until all structure is lost. The model then learns to denoise the perturbed images and restore their original structure through recursive applications of the noise prediction neural network $\epsilon_\theta ({x}t, t)$ over $T$ time-steps. In contrast to noise prediction, score-based generative models \cite{song2019generative} $\mathbf{s}{\boldsymbol{\theta}}({x})$ sample new instances by estimating the gradients of the log probability distribution of the data, $\nabla_{{x}} \log p_{\text {data }}({x})$. Diffusion models and score-based generative models can be formalized under a unified framework \cite{song2020score}.

Dhariwal et al. \cite{dhariwal2021diffusion} demonstrated that probabilistic diffusion models \cite{sohl2015deep} outperform prominent GANs \cite{goodfellow2014generative} in both quality and diversity metrics. Rombach et al. \cite{rombach2022high} introduced Stable Diffusion (SD), a groundbreaking approach that denoises images in latent space to reduce the computational and memory requirements associated with operating in pixel-level space. Stable Diffusion converts images to latent space via an image encoder and returns them to pixel space through an image decoder.

\subsection{Control Layout Overview} 
\label{Control-Layout-Overview} 
Conditional models \cite{yang2023reco, li2023gligen, wang2024instancediffusion, zhang2023adding, qin2023unicontrol, zhao2024uni, mou2024t2i} empower image generative models, particularly text-to-image models, to produce controllable outputs by extending them to accept a single or multiple external layout maps, commonly referred to as conditions. Scribbles and sketches are simple, vague types of conditions that can be easily created by users. In contrast, segmentation maps \cite{zhou2017scene} offer granular control over the image content by specifying a pixel-level condition map of the image lattice. Hough lines \cite{gu2022towards} and Canny edges \cite{canny1986computational} condition generative models to capture subtle textures and fine-grained details that cannot be specified through bounding boxes and segmentation maps. Keypoints \cite{cao2017realtime} provide a format for generating images based on skeletal structures, which likely depict human postures. Advanced control formats include normal maps \cite{vasiljevic2019diode} and depth maps \cite{ranftl2020towards}, both of which broaden the scope of the conditioning task by incorporating information about the perceived depth of images.

\subsection{Conditional Image Generative Models} T2I models utilize image captions to guide the generation process. Although simple and effective, text prompts alone are insufficient for tailoring the generation process to produce customized outputs. To address this limitation, various layout-to-image models \cite{jahn2021high, li2021image, sun2019image, sylvain2021object, yang2022modeling, zhao2019image} have been proposed, which employ more advanced conditional modalities, outlined in Section \ref{Control-Layout-Overview}, to enable fine-grained control. Conditional models generate complex images with style and content that comply with the provided control layout. PITI \cite{wang2022pretraining} is a layout-to-image diffusion model trained from scratch on conditional tokens. Training generative models from scratch is inefficient and incurs high training costs, as it does not benefit from the vast generation knowledge available in pretrained T2I models. To harness the pretraining knowledge, T2I-Adopters \cite{mou2024t2i} are lightweight neural network blocks adapted from NLP \cite{houlsby2019parameter} that facilitate the fine-tuning of large T2I models on various types of guiding inputs, such as depth, segmentation, keypoints, sketches, and color maps.

ReCo \cite{yang2023reco} fine-tunes a pretrained T2I model using bounding box and open-ended text information. However, this continuous learning process may result in a catastrophic loss of generation knowledge \cite{hu2021lora, ruiz2023dreambooth}. To mitigate this issue, GLIGEN \cite{li2023gligen} proposes freezing the weights of a pretrained SD model \cite{rombach2022high} and inserting trainable gated self-attention layers \cite{vaswani2017attention} into the attention blocks of the frozen SD model to handle the external grounding input. InstanceDiffusion \cite{wang2024instancediffusion} advances the grounding task further by introducing a model for precise object-level control while simultaneously enhancing image fidelity and multi-object sampling capabilities.

Zhang et al. \cite{zhang2023adding} goes further by introducing ControlNet \cite{zhang2023adding}, a groundbreaking general-purpose conditional model. They achieve this by freezing the weights of the pretrained model and training a copy of the pretrained locked backbone model. The locked model is preceded by a sequence of convolution layers that map the conditions into the trainable network. UniControl \cite{qin2023unicontrol} can simultaneously condition on an array of conditions by embedding all task instructions using a task-aware HyperNet \cite{ha2016hypernetworks, von2019continual}. UniControl demonstrates zero-shot conditioning capabilities for previously unseen control types. Uni-ControlNet builds on ControlNet by introducing a unified framework to simultaneously handle multiple control formats. Uni-ControlNet \cite{zhao2024uni} enables training and inference with a composite of modalities without the need for separate training for each modality.

\subsection{Training-Free Conditional Methods} Traditional conditional models require computationally expensive training on datasets of image-control pairs. Training-free methods \cite{fan2023frido, frolov2021attrlostgan, li2020bachgan, li2021image, sun2019image, yang2022modeling, zhao2019image} circumvent this by performing conditioning on pretrained T2I models during the sampling process, thereby eliminating the need for training or fine-tuning. BoxDiff \cite{xie2023boxdiff} achieves positional alignment by manipulating specific regions of the cross-attention map that correspond to constraining bounding box regions. Guidance-based methods can achieve control by designing task-specific guidance procedures. For example, Layout Control Guidance \cite{chen2024training} applies forward and backward guidance losses on the attention map to steer the sampling process towards generating object-level control. Universal Guidance \cite{bansal2023universal} offers a solution to avoid training a network on noisy data by leveraging available pretrained models, such as semantic segmentation and object recognition neural networks. The goal is to carefully design and minimize the loss function $\ell\left(c, f\left(\hat{z}_0\right)\right)$ between the guidance function $f(\cdot)$ of the latent image $z$ and the control module $c$.

Despite their success, training-free control approaches can prolong sampling time and decrease perceptual quality. As a result, they are typically used in conjunction with regular training-based conditional models.

\section{Method}

We introduce ObjectDiffusion, a model designed for grounded image generation based on object names and their locations. ObjectDiffusion draws inspiration from two notable conditional models, GLIGEN \cite{li2023gligen} and ControlNet \cite{zhang2023adding}, to seamlessly condition text-to-image diffusion models using a control layout. Specifically, we modify and combine the powerful network architecture of ControlNet with the condition processing and injection techniques proposed in GLIGEN. Our grounding layout consists of object detection annotations, where each object is defined by an open-ended text description \cite{yang2023reco}, rather than a label as in conventional object detection tasks, along with a bounding box that specifies the location of the object in the image. ObjectDiffusion incorporates the grounding information described above, along with a single image caption, to generate high-fidelity images where objects specified by the entity names are rendered in the regions outlined by the bounding boxes. In this section, we explain the control layout and break down the encoding techniques used to process and integrate the condition inputs into the baseline model. We also present the overall model architecture and articulate the learning objective.

\begin{figure}
    \centering
    \includegraphics[width=1\linewidth]{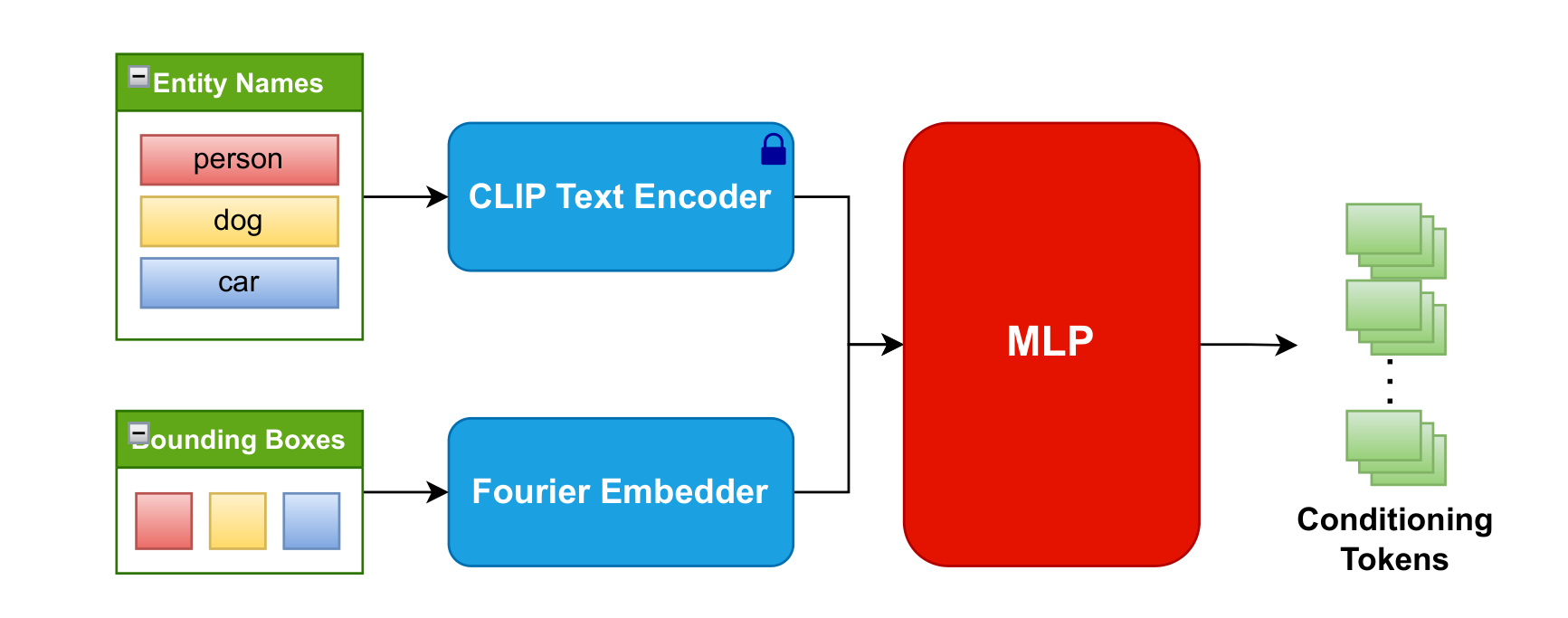}
    \caption{The grounding tokens are formed by fusing the CLIP \cite{radford2021learning} encoded object entities with their Fourier \cite{mildenhall10representing} embedded bounding boxes. The concatenation vector is processed through an MLP \cite{popescu2009multilayer}, which produces a standard-sized output vector of 768.}
    \label{fig:Processing-Grounding-Tokens}
\end{figure}

\subsection{Entity Grounding Layout} \label{sec:Entity-Grounding-Layout}
Due to its simplicity and effectiveness in controlling the generation process, we focus on grounding diffusion-based models on object detection annotations. The object detection serves as a condition to ground T2I models into rendering specific objects in specified regions of the synthesized images. In particular, our control layout consists of a set of grounding tokens, each of which is composed of one-to-one pairs of semantic and spatial information. The semantic instruction is a textual description of the object, for example, \textit{a shiny luxury red sports car}. Each object in the image can be  defined in an open-ended text format \cite{yang2023reco}, rather than being limited to a predefined set of labels like COCO categories \cite{lin2014microsoft}. For spatial instruction, each entity is attributed by a single 2D rectangular box which determines the location of the object in the image. The box is represented in the form of an object detection rectangular box defined by the two points, the upper-left point $(x_1,y_1)$ and the lower-right point $(x_2,y_2)$, needed to construct the detection box $[x_1, y_1, x_2, y_2]$. While the objects of the image are defined by the control layout, the overall image outline is specified by a single global image caption, similar to T2I models.

\begin{figure*}[!ht]
    \centering
    \includegraphics[width=1\linewidth]{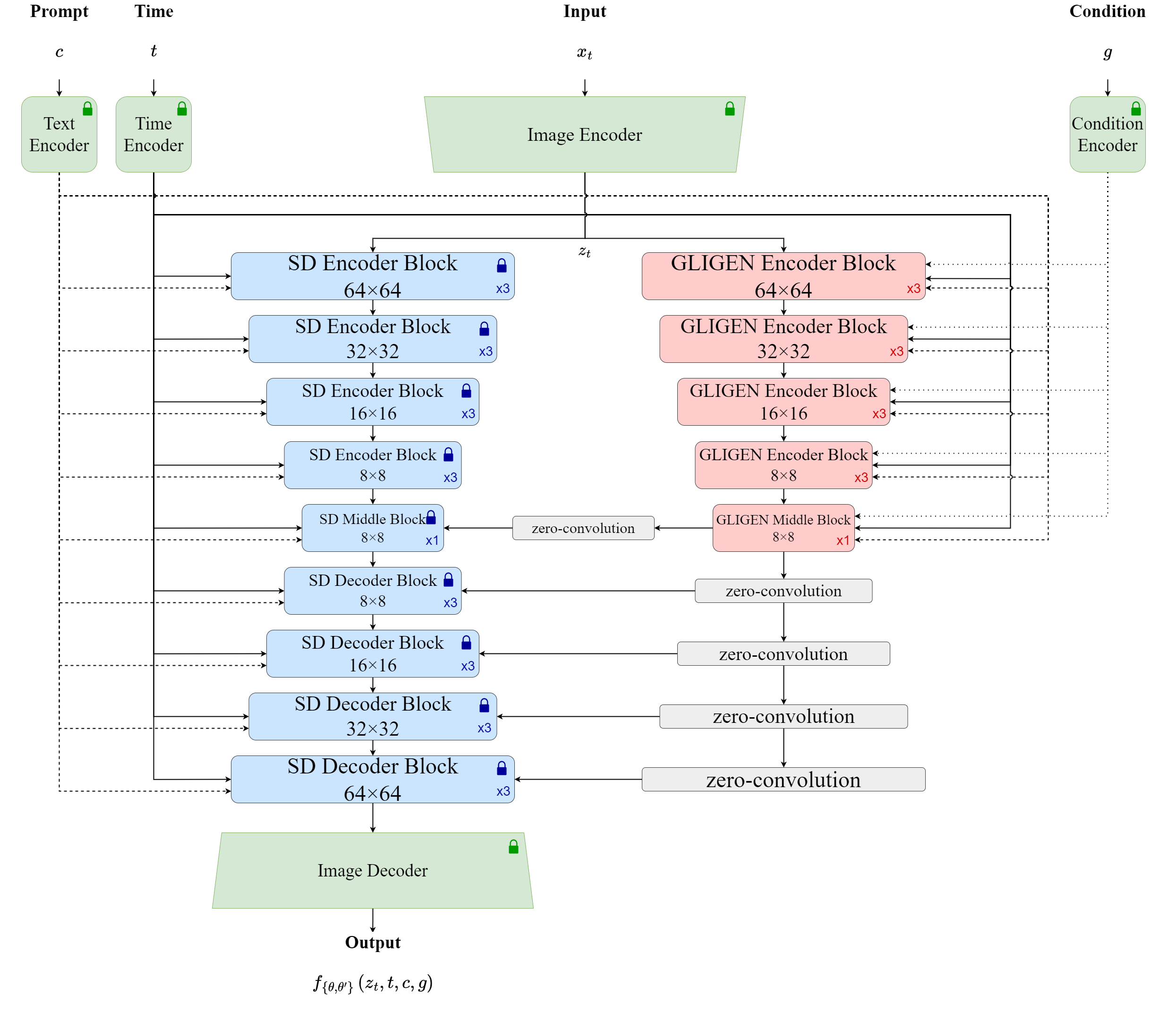}
    \caption{ObjectDiffusion architecture is divided into two parallel networks. For the first network, we utilize Stable Diffusion 1.4v \cite{rombach2022high} as a pretrained text-to-image model, depicted in blue. The second network is a trainable GroundNet, which consists of the encoder and middle blocks from GLIGEN \cite{li2023gligen}, represented in red. GroundNet injects the encoded conditional layout $g$, which consists of the positional and semantic tokens. During training, both networks receive the time $t$, the caption $c$, and the latent noised image input $z_t$. The two networks are connected via zero-convolution layers, highlighted in gray. ObjectDiffusion operates in latent space. The Image Encoder and Image Decoder project the input image from pixel space to latent space, and the output image from latent space back to pixel space, respectively. Our model architecture design is inspired by ControlNet \cite{zhang2023adding}.}
    \label{fig:ObjectDiffusion-architecture}
\end{figure*}

\subsection{Grounding Method Overview}
\label{sec:Grounding-Method-Overview}

\textbf{Processing Semantic Grounding Tokens:} Conditional semantic tokens are strings of varying lengths consisting of one or more words. These object-level text captions are processed similarly to regular image captions. Specifically, they are encoded using a pretrained CLIP ViT-L/14 \cite{radford2021learning} text encoder. The CLIP encoder, denoted as \( \mathcal{E}_{\text{text}}( e ) \), embeds the entity \( e \) into a tensor feature of length 768.

\textbf{Processing Spatial Grounding Tokens:} We embed the spatial information, presented as bounding boxes, into an expressive periodic tensor via a Fourier embedding \cite{mildenhall10representing} block. The mathematical representation of the Fourier transformation $F(\cdot)$ of a box $b = [x_1, y_1, x_2, y_2]$ is expressed as follows:
\begin{equation}
\begin{split}
F_i(b) &= \left[\sin \left(f_i \cdot x_1\right), \cos \left(f_i \cdot x_1\right), \ldots, \right. \\
\ldots, &\left. \sin \left(f_i \cdot y_2\right), \cos \left(f_i \cdot y_2\right)\right]
\end{split}
\end{equation}
\begin{equation}
\begin{split}
F(b) &= \left[F_0(b), F_1(b), \ldots, F_{M-1}(b)\right], 
\\
\text{where } &i \in \{0, 1, 2, \ldots, M-1\}
\end{split}
\end{equation}
\noindent where $M$ is the number of frequencies, and $F_i(b)$ is the Fourier transformation at the $f_i$ frequency. The output of the embedding $F(b)$ is 4 $\times$ 8 $\times$ 2 (box coordinates $\times$ frequency $\times$ sin \& cos), which results in a 64-dimensional tensor per bounding box. The Fourier embedder does not add any trainable parameters.

\textbf{Fusing Conditional Features:}
For each grounding entity, we first concatenate the encoded text entity $\mathcal{E}_{\text{text}}( e )$ with the embedded positional information $F(b)$ into a single tensor of size 832. Then, we feed this tensor as input to a Multilayer Perceptron (MLP) \cite{popescu2009multilayer} consisting of three fully connected linear layers of sizes (832, 512, 768) and a Sigmoid-weighted Linear Unit (SiLU) \cite{elfwing2018sigmoid}:
\begin{gather}
o_i=\operatorname{MLP}\left(\mathcal{E}_{\text{text}}( e ), \operatorname{F}({b})\right) \\
o=\left[o_1, o_2, \ldots, o_{N}\right], \quad \text { where } i \in\{1,2, \ldots, N\} 
\end{gather}
\noindent The MLP transforms the concatenated grounding feature vectors from 832 to a transformer vector $o_i$ of standard size 768. Assuming that the maximum number of entities allowed per image is a hyperparameter $N$, the final control vector $o$ has size ($N$, 768). Figure \ref{fig:Processing-Grounding-Tokens} illustrates the processing and fusion of the grounding inputs.

\textbf{Integrating Conditional Features:}
To inject the extracted control features $o$ into the standard transformer blocks of LDM \cite{rombach2022high}, we follow GLIGEN and insert a new gated self-attention $\operatorname{SA}\left(\cdot\right)$ layer between the self-attention and the cross-attention of the transformer block \cite{alayrac2022flamingo}. The gated self-attention layer takes both the intermediate latent visual features $v$ and the external control features $o$ as inputs and integrates them, as shown in the equation:
\begin{equation}
v=v+ \tanh (\gamma) \cdot \operatorname{SA}\left(\left[v, o\right]\right)
\end{equation}

\subsection{Model Architecture}
\label{sec:Model-Architecture}
Our model architecture is illustrated in Figure \ref{fig:ObjectDiffusion-architecture}. We begin with a pretrained Stable Diffusion (SD) model \cite{rombach2022high}, which enables conditioning image generative diffusion models on captions but does not support conditional layouts, such as the object detection annotations defined in Section \ref{sec:Entity-Grounding-Layout}. We lock the entire SD model and clone the encoder and middle blocks, but not the decoder blocks, from the pretrained SD into a trainable network \cite{zhang2023adding}. This approach allows us to leverage the generation knowledge acquired from training on large-scale datasets. To endow our model with grounding capabilities, we introduce several modifications to the copied SD network structure. Specifically, we insert gated self-attention (GSA) layers, where gating is implemented via $\tanh (\gamma)$ \cite{dubey2022activation}, into each of the attention blocks of the trainable network \cite{li2023gligen}, as discussed in Section \ref{sec:Grounding-Method-Overview}. We refer to this modified trainable network as GroundNet, as it enables ObjectDiffusion to be grounded on the detection layout by injecting the grounding features into the pretrained structure. GroundNet is connected to the SD baseline through zero-convolution layers \cite{nichol2021improved}, implemented as 1 $\times$ 1 zero-initialized convolution layers. Both the gating mechanism of the GSAs and the zero-convolution layers are crucial for preventing noise, introduced by the new grounding features, from compromising the weights of the trainable GroundNet. This is particularly important during the initial training phase, where noise may disrupt the pretraining generation knowledge.

ObjectDiffusion is similar to ControlNet in that both freeze the pretrained SD model and fine-tune a parallel network to adapt to a new control layout. Additionally, both use zero-convolution layers to connect the trainable network with the frozen baseline network. However, ObjectDiffusion and ControlNet differ in several key aspects. First, ControlNet fine-tunes an exact copy of the locked SD blocks, whereas ObjectDiffusion modifies the cloned SD blocks by incorporating a gated self-attention mechanism that enables the integration of grounding features. Second, ControlNet adds the extracted conditional features to the visual features and injects the concatenated features into only the first trainable encoder of the U-Net \cite{ronneberger2015u}. In contrast, ObjectDiffusion implements a multi-scale control injection, where the control features are injected into each of the encoder layers and the middle layer of GroundNet. The multi-scale control injection strategy amplifies the control signal and has been shown to produce more precise outputs compared to single injection \cite{zhao2024uni}. Finally, ControlNet processes the control input using an auxiliary convolution network. In contrast, ObjectDiffusion processes the conditional input via the CLIP text encoder and Fourier Embedder, as presented in Section \ref{sec:Grounding-Method-Overview}. It is worth noting that the object detection layout outlined in Section \ref{sec:Entity-Grounding-Layout} is not among the control layouts supported by the original ControlNet.

\subsection{Learning Objective}
Diffusion models \cite{sohl2015deep} learn the distribution of the training data by maximizing the log-likelihood. Given a noised image ${x}_t$, which is equivalent to ${z}_t$ in latent space, and the number of steps $t$ at which the ground truth image ${x}_0$ was distorted, a diffusion neural network is trained to predict the amount of noise ${\epsilon}$ added to the original image \cite{ho2020denoising}. Accordingly, the training objective is a weighted variational bound that can be interpreted as a simple L2 loss function between the noise added during the diffusion process and the noise predicted by the model:
\begin{equation}
\min _{{\theta}} \mathcal{L}_{\mathrm{DM}}=\mathbb{E}_{{x}, t, {\epsilon} \sim \mathcal{N}({0}, \mathrm{I})}\left[\left\|{\epsilon}-f_{{\theta}}\left({x}_t, t\right)\right\|_2^2\right] 
\end{equation}
where $\epsilon$ is the ground-truth noise, $\theta$ is the learnable model parameters, and $f_{\theta}$ is the predicted noise by the network.

We condition the latent diffusion model $f_{\theta}\left({z}_t, t\right)$ on two external inputs: the text caption $c$ and the object grounding input $g$, expressed as:
\begin{equation}
g =\left[\left(e_1,  {b}_1\right), \cdots,\left(e_N, {b}_N\right)\right]
\end{equation}
where $e_i$ and $b_i$ represent the semantic and positional information, respectively, corresponding to the $i\textsuperscript{th}$ grounding object.

As discussed in Section \ref{sec:Model-Architecture}, our proposed model consists of two networks. The first is a frozen T2I diffusion network $f_{\theta}\left({z_t}, t, c\right)$ \cite{rombach2022high}, which conditions the model on the text caption $c$ but not on the grounding information $g$. The second is a trainable GroundNet $f_{\theta^{\prime}}\left(z_t, t, c, g\right)$, which additionally conditions on both the caption $c$ and the grounding information $g$. All components of GroundNet are trainable. The overall noise prediction of our model can be formalized as follows:
\begin{equation}
f_{\left\{\theta, \theta^{\prime}\right\}}\left(z_t, t,c, g\right) = f_{\theta}\left(z_t, t, c\right) +\mathnormal{Z}\left(f_{\theta^{\prime}}\left(z_t, t, c,  g\right)\right)
\end{equation}
where $\mathnormal{Z}$ denotes the zero-convolution layers. For simplicity, we omit the weights of $\mathnormal{Z}$ and assume that the weights of the zero-convolution layers are incorporated within the weights ${\theta^{\prime}}$.

Our learning objective minimizes the L2 loss function, which quantifies the difference between the noise $\epsilon$ added to the latent image during the forward process and the noise predicted by the model $f_{\left\{{\theta}, {\theta^{\prime}}\right\}}\left({z}_t, t,c,g\right)$: 
\begin{equation}
\min _{{\theta^{\prime}}} \mathcal{L}=\mathbb{E}_{{z}, t, c, g,{\epsilon} \sim \mathcal{N}({0}, \mathrm{I})}\left[\left\|{\epsilon}-f_{\left\{{\theta}, {\theta^{\prime}}\right\}}\left(\mathrm{z}_t, t,c,g\right)\right\|_2^2\right]
\end{equation}

\section{Experiments}
\subsection{Experimental Setup}

\textbf{Dataset:}
We fine-tune and evaluate our model on the Common Objects in Context (COCO) dataset \cite{lin2014microsoft}. We use the COCO2017 training split, which contains 118k annotated images, for the purpose of fine-tuning, and the COCO2017 validation split, consisting of 5k annotated images, for evaluation. The median image resolution in COCO2017 dataset is 640 $\times$ 480 pixels, with each image containing five captions and an average of seven object labels along with their corresponding bounding boxes. Objects are classified into 80 simplified categories.

\begin{figure}
    \centering
    \includegraphics[width=1\linewidth]{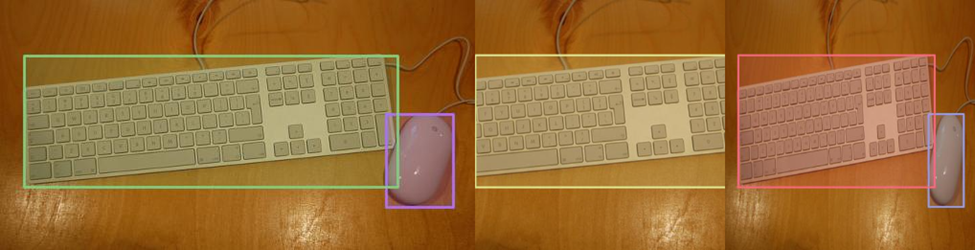}
    \caption{This figure highlights the difference between the resizing via Bicubic \cite{keys1981cubic} interpolation that we apply and the center cropping implemented in GLIGEN. The original image is on the left, the GLIGEN \cite{li2023gligen} preprocessed image is in the middle, and our preprocessed image is on the right.}
    \label{fig:Preprocessing-exmaple-1}
\end{figure}

\textbf{Preprocessing Pipeline:}
We resize the training images using Bicubic interpolation \cite{keys1981cubic} to 512 $\times$ 512 pixels. Unlike GLIGEN, we do not crop the image at all. Figure \ref{fig:Preprocessing-exmaple-1} illustrates the difference between our image preprocessing and the preprocessing pipeline applied by GLIGEN. We randomly flip the image 50\% of the time and discard the object grounding annotations when the area of their bounding boxes is less than 1\% of the total image area. Additionally, we retain only the 30 objects with the largest areas. The bounding boxes are converted from the COCO format $[x_{min}, y_{min}, w, h]$ to the format $[x_{min}, y_{min}, x_{max}, y_{max}]$.

\textbf{Initialization:}
Initializing our model weights from pretrained models allows us to leverage and build upon the extensive knowledge acquired through pretraining. We initialize the frozen T2I model using Stable-Diffusion-v-1-4 \cite{sdchpt1-4v}, which is extensively trained on a high-resolution subset of the large-scale LAION-5B \cite{schuhmann2022laion} dataset. We initialize the trainable GroundNet using the GLIGEN diffusers-generation-text-box checkpoint \cite{gligenchpt}, which is pretrained on manually annotated object names and bounding boxes from the Object365 \cite{shao2019objects365}, Flickr \cite{plummer2015flickr30k}, and Visual Genome (VG) \cite{krishna2017visual} datasets, as well as object names and bounding boxes extracted via GLIP \cite{li2022grounded} from the SBU \cite{ordonez2011im2text} and CC3M \cite{sharma2018conceptual} datasets, but not fine-tuned on the COCO \cite{lin2014microsoft} dataset. We freeze the pretrained CLIP ViT-L/14 \cite{radford2021learning} text encoder for the entire training process. ObjectDiffusion contains 1.32B parameters, of which 460.8M are trainable and 860M are frozen.

\textbf{Implementation Details:}
We fine-tune ObjectDiffusion for 100k iterations on a single NVIDIA QUADRO GV100 32 GB GPU using the Adam optimizer \cite{kingma2014adam} with a learning rate of 5e-5. We set the warmup iterations to 4k and apply no weight decay. We randomly select one out of the five available COCO captions. To ensure robust training, we randomly drop the caption and randomly drop the condition input, each with a 10\% independent probability.

We utilize several techniques to mitigate the constraints of GPU memory and computational capacity. First, we implement Automatic Mixed Precision (AMP) \cite{micikevicius2017mixed}, a smart quantization technique that assigns multiple precisions to different modules of the same network based on their tolerance of low precision. AMP delivers a 109\% speedup in training. Second, we employ Gradient Accumulation \cite{kozodoi2021}, where we train with a batch size of 4 but update the model weights with the normalized accumulated gradients once every 4 iterations. Gradient Accumulation allows us to quadruple the batch size from 4 to 16, which smooths the learning curve.

\textbf{Evaluation Benchmarks:}
We calculate the Average Precision (AP) \cite{padilla2020survey} and the Average Recall (AR) \cite{padilla2020survey} on 5k generated images using a pretrained YOLOv8m \cite{redmon2016you} to measure the precision of our model in rendering the correct objects in the intended locations. We quantify image quality by generating 20k images and computing the Fréchet Inception Distance (FID) \cite{heusel2017gans} score between the generated images and real images. All images used in the closed-set evaluation are generated from the COCO2017 evaluation set annotations. For the open-set assessment, we craft image captions and grounding inputs and qualitatively analyze the output visuals.

\begin{figure}
    \centering
    \includegraphics[width=1\linewidth]{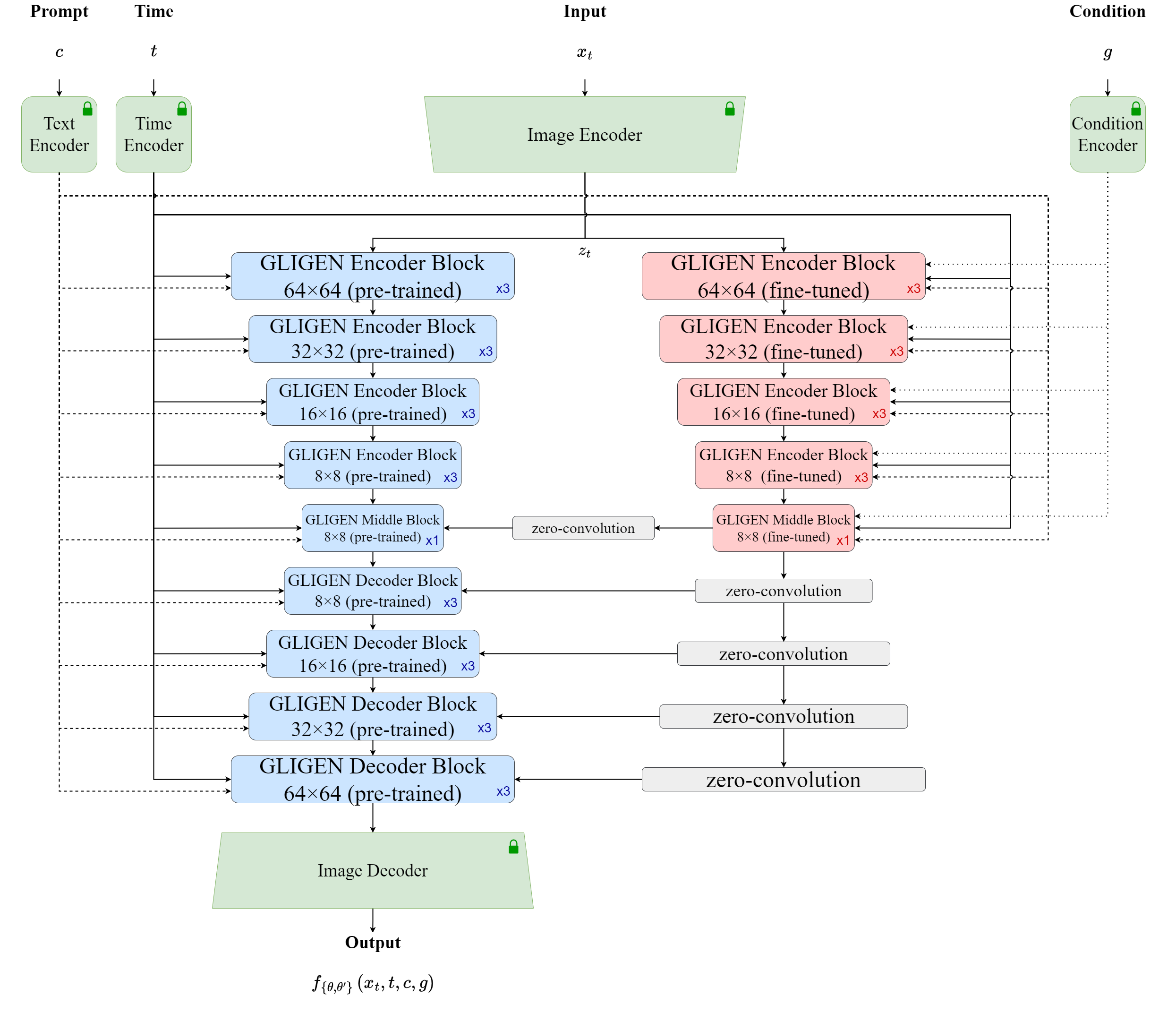}
    \caption{The inference schema consists of two networks. The first network, displayed in red, is our GroundNet fine-tuned on COCO2017 \cite{lin2014microsoft} object detection annotations. The second network, displayed in blue, is a pretrained GLIGEN \cite{li2023gligen}. We replace the pretrained Stable Diffusion \cite{rombach2022high} model used during training with the pretrained GLIGEN because it yields more precise grounding abilities. GLIGEN is trained on the Object365 \cite{shao2019objects365}, GoldG \cite{li2022grounded} (Flickr and VG), SBU \cite{ordonez2011im2text}, and CC3M \cite{sharma2018conceptual} datasets, but not on the COCO dataset.}
    \label{fig:ObjectDiffusion-architecture-inference}
\end{figure}

\textbf{Inference:}
During inference, we retain the fine-tuned GroundNet and replace the SD model with a pretrained GLIGEN network, as we have empirically verified that GLIGEN yields more controllable images. The inference model architecture is depicted in Figure \ref{fig:ObjectDiffusion-architecture-inference}. In all of our experiments, we sample 512 $\times$ 512 RGB images using the Pseudo Linear Multi-Step (PLMS) \cite{liu2022pseudo} sampler with 50 steps and a guidance scale of 7.5.

 \begin{table*}[htbp]
    \centering
    \begin{tabular}{l|ccc}
        \hline 
        \multirow{2}{*}{Model} & \multicolumn{3}{c}{YOLOv8m \cite{redmon2016you}} \\
        & AP (\(\uparrow\)) & AP$_{\text{50}}$ (\(\uparrow\)) & AP$_{\text{75}}$ (\(\uparrow\)) \\
        \hline upper-bound  & 49.7 & 65.5 & - \\
        \hline LostGAN-V2 \cite{sun2021learning}  & 9.1 & 15.3 & 9.8 \\
        LAMA \cite{li2021image} & 13.40 & 19.70 & 14.90 \\
        TwFA \cite{yang2022modeling} & - & 28.20 & 20.12 \\
        GLIGEN \cite{li2023gligen} & 22.4 & 36.5 & 24.1 \\
        \hline \multicolumn{1}{c}{ pretrained on (Object365, GoldG, SBU, CC3M) } \\
        GLIGEN (zero-shot) \cite{li2023gligen} & 19.1 & 30.5 & 20.8 \\
        GLIGEN (fine-tuned) \cite{li2023gligen} & $\mathbf{3 0 . 8}$ & 42.3 & $\mathbf{35.3}$ \\
        ObjectDiffusion (ours)& 27.4  & $\mathbf{46.6}$ & 28.2 \\
        \hline
    \end{tabular}
    \caption{ObjectDiffusion outperforms the SOTA model on AP$_{\text{50}}$, indicating its ability to render objects with high precision. The evaluation of grounding precision on synthesized images is conducted using the captions, boxes, and object labels from the COCO2017 \cite{lin2014microsoft} validation set. The AP, AP$_{\text{50}}$, and AP$_{\text{75}}$ metrics \cite{padilla2020survey} for ObjectDiffusion and the upper bound are calculated using a pretrained YOLOv8m \cite{redmon2016you} object detection model on images of size 512 $\times$ 512. Other values are taken from \cite{li2023gligen}.}
    \label{tab:AP_scores}
\end{table*}

\subsection{Quantitative Results}
Table \ref{tab:AP_scores} details the Average Precision (AP, AP$_{\text{50}}$, AP$_{\text{75}}$) \cite{padilla2020survey} metrics and Average Recall (AR) \cite{padilla2020survey} metric between the YOLOv8m \cite{redmon2016you} predictions on synthesized images and ground truth detection annotations. ObjectDiffusion achieves 27.4, 46.6, 28.2, and 44.5 in AP, AP$_{\text{50}}$, AP$_{\text{75}}$, and AR, respectively.

Our model surpasses the state-of-the-art GLIGEN (fine-tuned) \cite{li2023gligen} model in the AP$_{\text{50}}$ score by more than 10\% and ranks second in AP and AP$_{\text{75}}$ scores, surpassing LostGAN-V2 \cite{sun2021learning}, LAMA \cite{li2021image}, TwFA \cite{yang2022modeling}, and GLIGEN \cite{li2023gligen}. Our high AP scores indicate that ObjectDiffusion has succeeded in learning from grounding information and is capable of synthesizing images in high alignment with the control layout. Our model increased the zero-shot AP score of GLIGEN (zero-shot) from 19.1 to 27.4 and the AP$_{\text{50}}$ score from 30.5 to 46.6, demonstrating successful fine-tuning on the COCO2017 \cite{lin2014microsoft} dataset. The results also show that our approach circumvents the problem of knowledge forgetting that typically hinders the fine-tuning of layout-to-image models.

In Table \ref{tab:AP_scores_class}, we calculate the AP and AP$_{\text{50}}$ score values per class and report the top three COCO classes with the highest AP values, as well as the top three COCO classes with the lowest AP values. The classes cat, bear, and dog are the top classes where our model grounded objects with very high precision. In contrast, the classes book, hair drier, and sports ball are the most confounding for our model. Unexpectedly, the effectiveness of grounding an object category does not correlate with its frequency in the fine-tuning dataset. For example, the category book is overrepresented in the COCO2017 dataset, yet it has a very low AP value. Similarly, the categories cat, dog, and bear are underrepresented in the COCO2017 dataset but have the highest AP values.

Table \ref{tab:AR_scores} shows that our model achieves the highest reported AR score of 44.5, which is only five points behind the upper bound value. The high AR value suggests that ObjectDiffusion rarely overlooks objects, a common issue that text-to-image models suffer from.

\begin{table}[htbp]
    \begin{tabular}{l|c|cc}
         \hline & \multirow{2}{*}{Class} & \multicolumn{2}{c}{YOLOv8m \cite{redmon2016you}} \\
         & & AP (\(\uparrow\)) & AP$_{\text{50}}$ (\(\uparrow\)) \\
        \hline 
        \multirow{3}{9em}{Highest AP Classes}  & 
        cat  & 74.2 & 96.7 \\
        & bear & 70.3 & 91.0  \\
        & dog & 62.7 & 87.5  \\
        \hline 
        \multirow{3}{9em}{Lowest AP Classes}  
        & book & 4.6  & 11.0 \\
        & hair drier & 4.3 & 6.1 \\
        & sports ball & 3.7  & 10.9 \\
        \hline
    \end{tabular}
    \caption{This table lists the COCO \cite{lin2014microsoft} classes with the highest AP scores (cat, bear, dog) and the COCO classes with the lowest AP \cite{padilla2020survey} scores (book, hair dryer, sports ball). The values are calculated using the YOLOv8m \cite{redmon2016you} object detection model on images synthesized from COCO2017 validation set annotations with an image size of 512 $\times$ 512.}
    \label{tab:AP_scores_class}
\end{table}

Finally, to measure the quality of our model generation, we calculate the FID \cite{heusel2017gans} scores in Table \ref{tab:FID_scores}. ObjectDiffusion achieves a FID score of 19.8, outperforming the state-of-the-art GLIGEN (fine-tuned), GLIGEN, and GLIGEN (zero-shot) models by 6\%, 8\%, and 27\%, respectively. The improvement in the FID score demonstrates the efficiency of our proposed training architecture in learning the generation capabilities for high-quality image synthesis.

It is worth noting that our evaluation is limited to models trained on publicly available datasets.

\subsection{Qualitative Results}
We visually demonstrate the capabilities of ObjectDiffusion in grounding on boxes and entity names. We test our model under both closed-set and open-set settings. Our qualitative testing is designed to encompass various entities, such as animals, persons, vehicles, household items, and sporting equipment, in both indoor and outdoor contexts. To verify the ability of our model to generate satisfactory images from challenging requirements, we vary the bounding box sizes, locations, the number of objects per image, and the length of the semantic descriptions of objects.

\begin{figure*}[htbp]

\centering
\begin{minipage}[t]{0.327\textwidth}
  \centering
  \captionsetup{font=small} 
\includegraphics[width=1\textwidth]{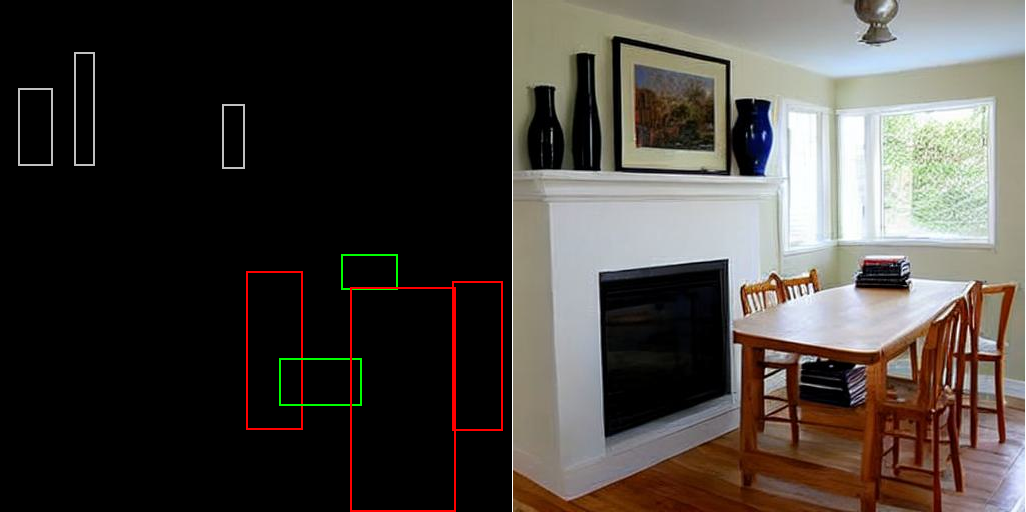}
\subcaption[first caption.]{
\underline{Image Caption:} “A room with a fireplace and a wooden table" \\\underline{Conditional Entities:} \textcolor{myred}{chair}, \textcolor{mygreen}{book}, \textcolor{mysilver}{vase}}\label{fig:1a}
\end{minipage}%
\hspace{0.001\textwidth}
\begin{minipage}[t]{0.327\textwidth}
  \centering
  \captionsetup{font=small} 
\includegraphics[width=1\textwidth]{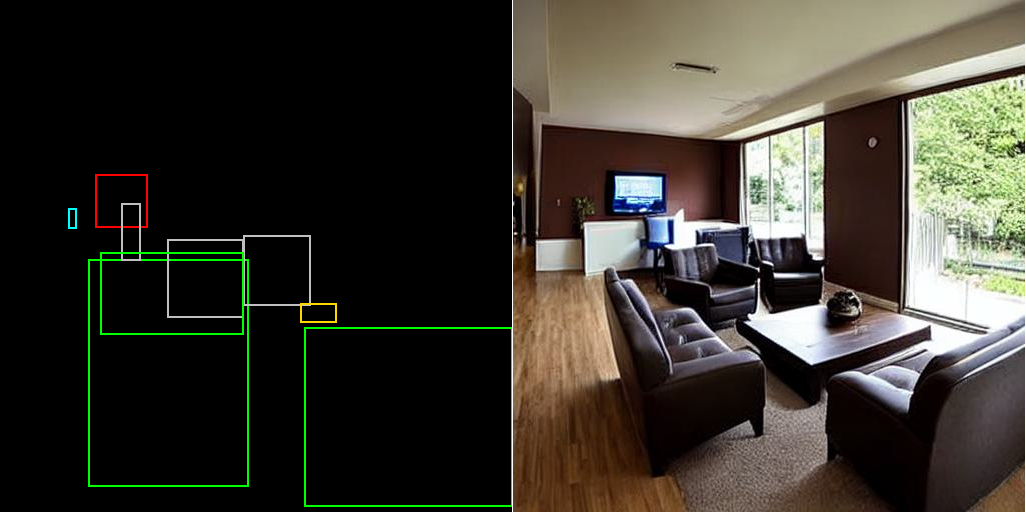}
\subcaption[second caption.]{
\underline{Image Caption:} “A living room that has a bunch of different couches" \\\underline{Conditional Entities:} \textcolor{myred}{tv}, \textcolor{mygreen}{couch}, \textcolor{mysilver}{chair}, \textcolor{mycyan}{vase}, \textcolor{mygold}{bowl}
}\label{fig:1b}
\end{minipage}%
\hspace{0.001\textwidth}
\begin{minipage}[t]{0.327\textwidth}
  \centering
   \captionsetup{font=small} 
\includegraphics[width=1\textwidth]{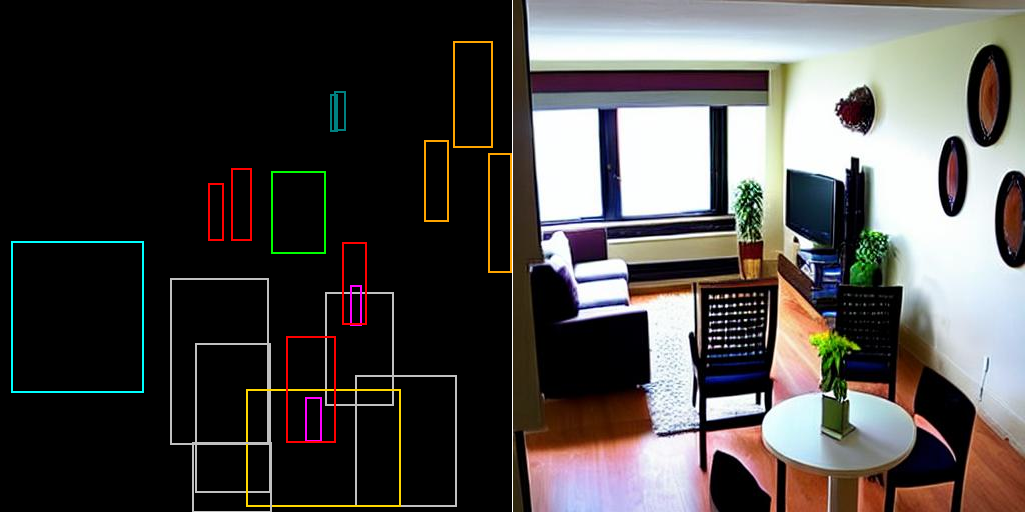}
\subcaption[third caption.]{
\underline{Image Caption:} “a room that has some furniture and a table in it" \\\underline{Conditional Entities:} \textcolor{myred}{potted plant}, \textcolor{mygreen}{tv}, \textcolor{mysilver}{chair}, \textcolor{mycyan}{couch}, \textcolor{mygold}{dining table}, \textcolor{mymagenta}{vase}, \textcolor{myteal}{bottle}, \textcolor{myorange}{bowl}
}\label{fig:1c}
\end{minipage}

\vspace{0.3cm}

\centering
\begin{minipage}[t]{0.327\textwidth}
  \centering
  \captionsetup{font=small} 
\includegraphics[width=1\textwidth]{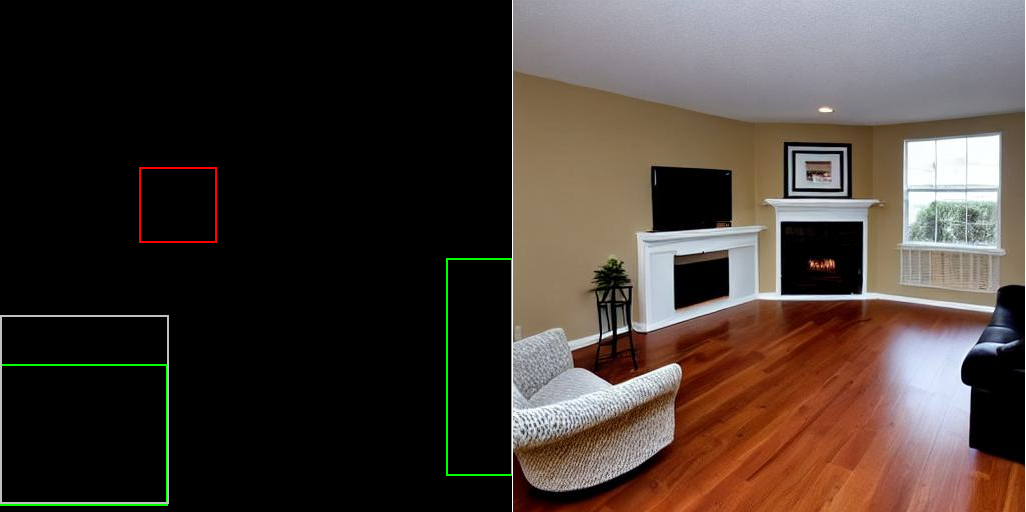}
\subcaption[first caption.]{
\underline{Image Caption:} “room with wood floors and a stone fire place" \\\underline{Conditional Entities:} \textcolor{myred}{tv}, \textcolor{mygreen}{couch}, \textcolor{mysilver}{chair}
}\label{fig:1a}
\end{minipage}%
\hspace{0.001\textwidth}
\begin{minipage}[t]{0.327\textwidth}
  \centering
  \captionsetup{font=small} 
\includegraphics[width=1\textwidth]{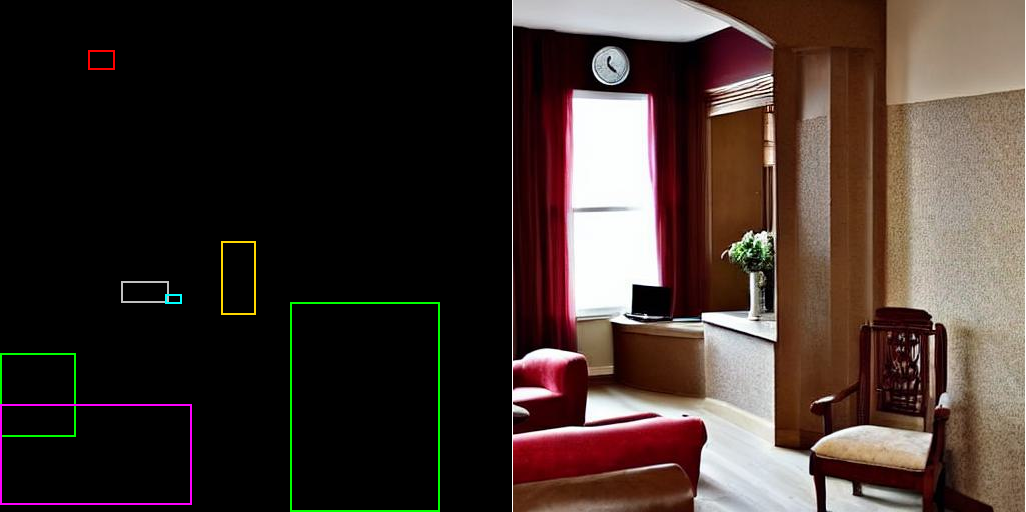}
\subcaption[second caption.]{
\underline{Image Caption:} “A beige living room with a cabinet,flowers, lamp and armchair" \\\underline{Conditional Entities:} \textcolor{myred}{clock}, \textcolor{mygreen}{chair}, \textcolor{mysilver}{laptop}, \textcolor{mycyan}{mouse}, \textcolor{mygold}{vase}, \textcolor{mymagenta}{couch}
}\label{fig:1b}
\end{minipage}%
\hspace{0.001\textwidth}
\begin{minipage}[t]{0.327\textwidth}
  \centering
   \captionsetup{font=small} 
\includegraphics[width=1\textwidth]{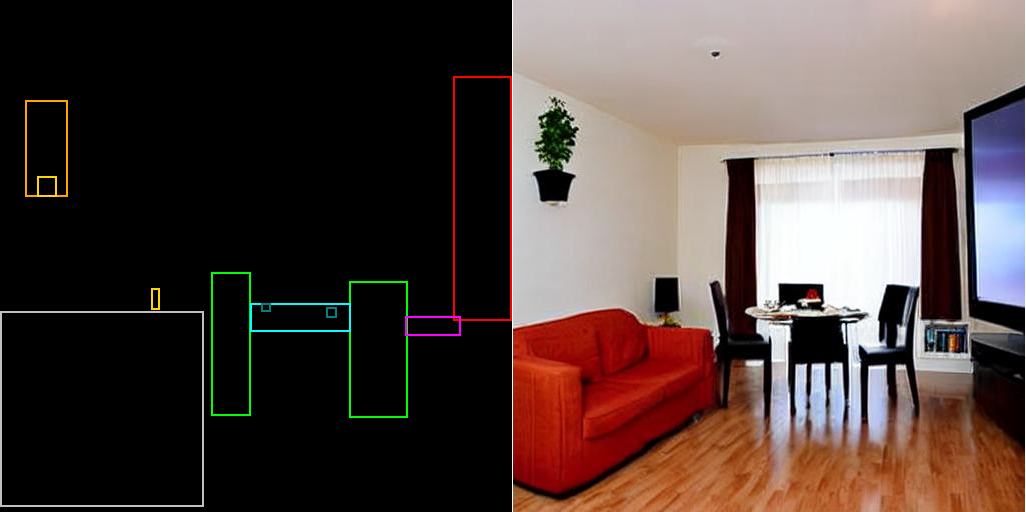}
\subcaption[third caption.]{
  \underline{Image Caption:} “A room with a couch, table set with dinnerware and a television" \\\underline{Conditional Entities:} \textcolor{myred}{tv}, \textcolor{mygreen}{chair}, \textcolor{mysilver}{couch}, \textcolor{mycyan}{dining table}, \textcolor{mygold}{vase}, \textcolor{mymagenta}{book}, \textcolor{myteal}{cup}, \textcolor{myorange}{potted plant}
}\label{fig:1c}
\end{minipage}

\vspace{0.3cm}

\centering
\begin{minipage}[t]{0.327\textwidth}
  \centering
  \captionsetup{font=small} 
\includegraphics[width=1\textwidth]{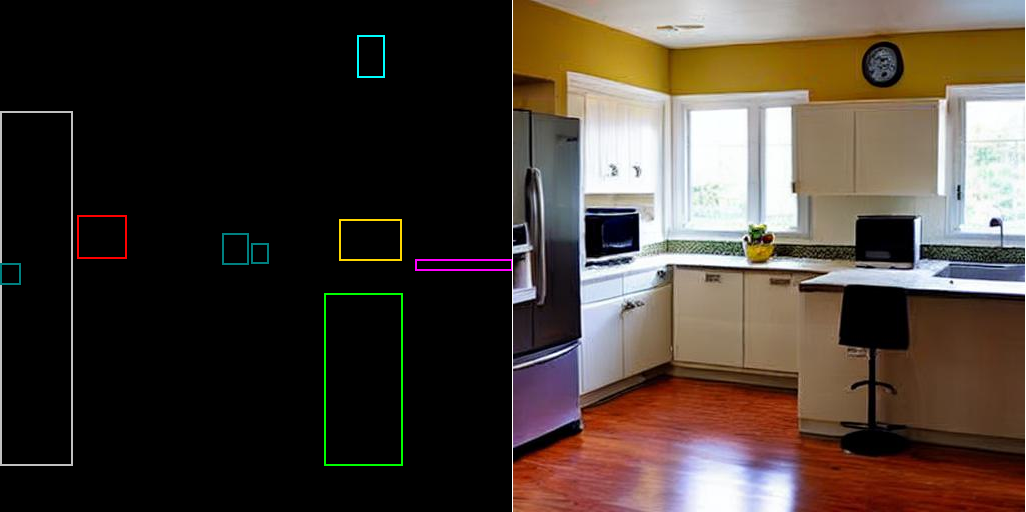}
\subcaption[first caption.]{
\underline{Image Caption:} “A kitchen that has carpeted floors and wooden cabinets" \\\underline{Conditional Entities:} \textcolor{myred}{tv}, \textcolor{mygreen}{chair}, \textcolor{mysilver}{refrigerator}, \textcolor{mycyan}{clock}, \textcolor{mygold}{microwave}, \textcolor{mymagenta}{sink}, \textcolor{myteal}{bowl}
}\label{fig:1a}
\end{minipage}%
\hspace{0.001\textwidth}
\begin{minipage}[t]{0.327\textwidth}
  \centering
  \captionsetup{font=small} 
\includegraphics[width=1\textwidth]{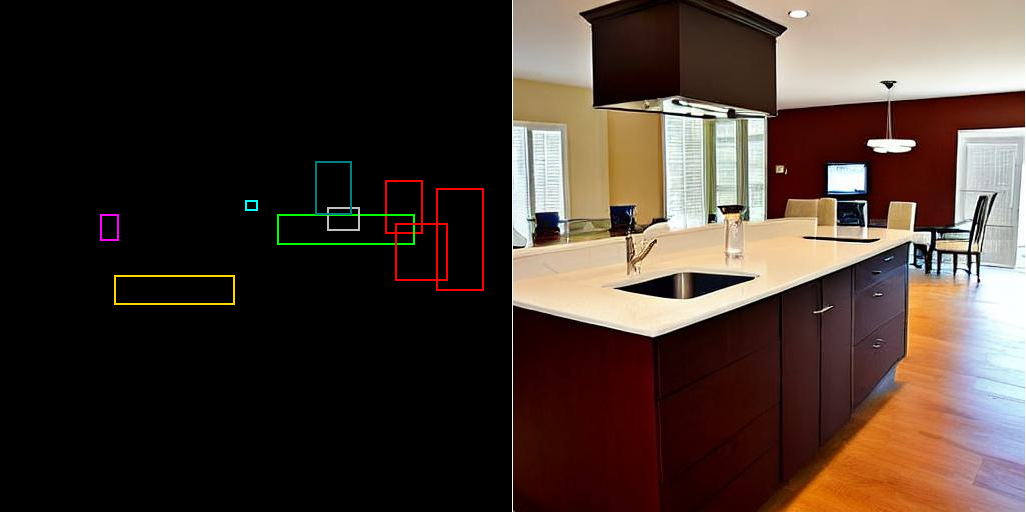}
\subcaption[second caption.]{
 \underline{Image Caption:} “A kitchen with a counter and a table with chairs" \\\underline{Conditional Entities:} \textcolor{myred}{chair}, \textcolor{mygreen}{dining table}, \textcolor{mysilver}{bowl}, \textcolor{mycyan}{apple}, \textcolor{mygold}{sink}, \textcolor{mymagenta}{cup}, \textcolor{myteal}{tv}
}\label{fig:1b}
\end{minipage}%
\hspace{0.001\textwidth}
\begin{minipage}[t]{0.327\textwidth}
  \centering
   \captionsetup{font=small} 
\includegraphics[width=1\textwidth]{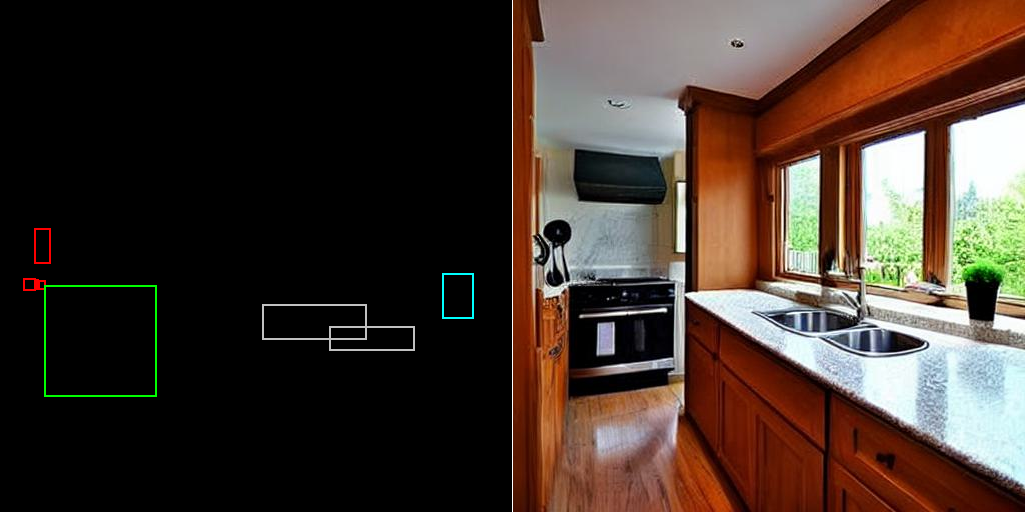}
\subcaption[third caption.]{
\underline{Image Caption:} “Two sinks that are in a kitchen near a window" \\\underline{Conditional Entities:} \textcolor{myred}{spoon}, \textcolor{mygreen}{oven}, \textcolor{mysilver}{sink}, \textcolor{mycyan}{potted plant}
}\label{fig:1c}
\end{minipage}

\vspace{0.3cm}

\centering
\begin{minipage}[t]{0.327\textwidth}
  \centering
  \captionsetup{font=small} 
\includegraphics[width=1\textwidth]{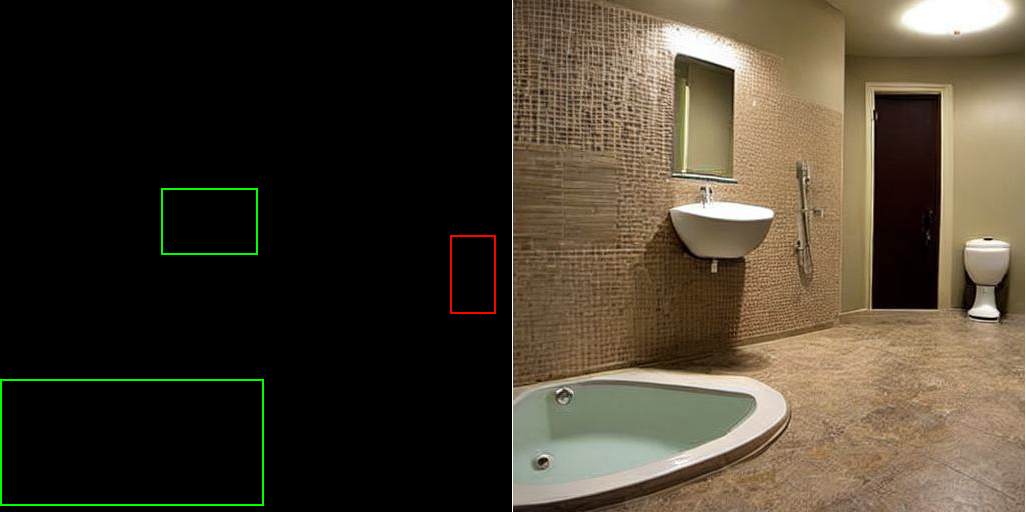}
\subcaption[first caption.]{
\underline{Image Caption:} “a fancy bathroom with some tile walls" \\\underline{Conditional Entities:} \textcolor{myred}{toilet}, \textcolor{mygreen}{sink}
}\label{fig:1a}
\end{minipage}%
\hspace{0.001\textwidth}
\begin{minipage}[t]{0.327\textwidth}
  \centering
  \captionsetup{font=small} 
\includegraphics[width=1\textwidth]{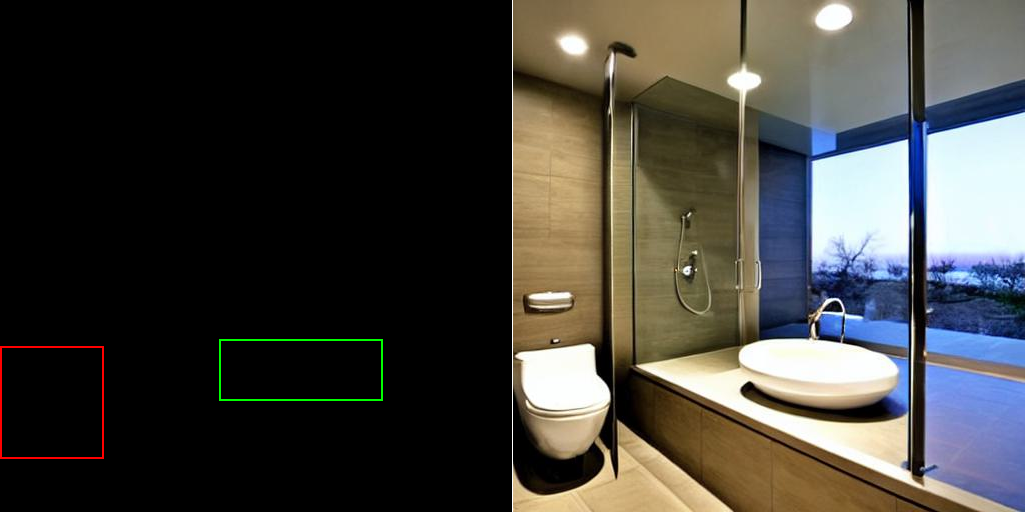}
\subcaption[second caption.]{
\underline{Image Caption:} “A residential bathroom has a modern motel look with upgraded fixtures and accessories" \\\underline{Conditional Entities:} \textcolor{myred}{toilet}, \textcolor{mygreen}{sink}
}\label{fig:1b}
\end{minipage}%
\hspace{0.001\textwidth}
\begin{minipage}[t]{0.327\textwidth}
  \centering
   \captionsetup{font=small} 
\includegraphics[width=1\textwidth]{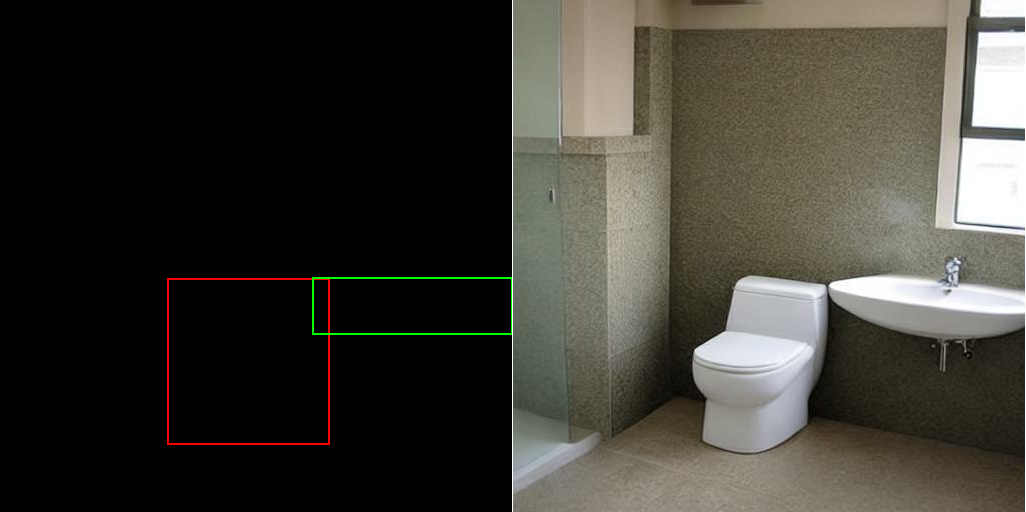}
\subcaption[third caption.]{
\underline{Image Caption:} “A view of a bathroom that is clean" \\\underline{Conditional Entities:} \textcolor{myred}{toilet}, \textcolor{mygreen}{sink}
}\label{fig:1c}
\end{minipage}

\caption{This figure represents qualitative examples of closed-set configuration. Our model demonstrates precise generation capabilities for complex indoor visuals, such as living rooms, kitchens, and bathrooms. Images are generated from COCO2017 \cite{lin2014microsoft} validation set annotations.}
\label{fig:closed-set-indoor}
\end{figure*}

\begin{figure*}[htbp]

\centering
\begin{minipage}[t]{0.327\textwidth}
  \centering
  \captionsetup{font=small} 
\includegraphics[width=1\textwidth]{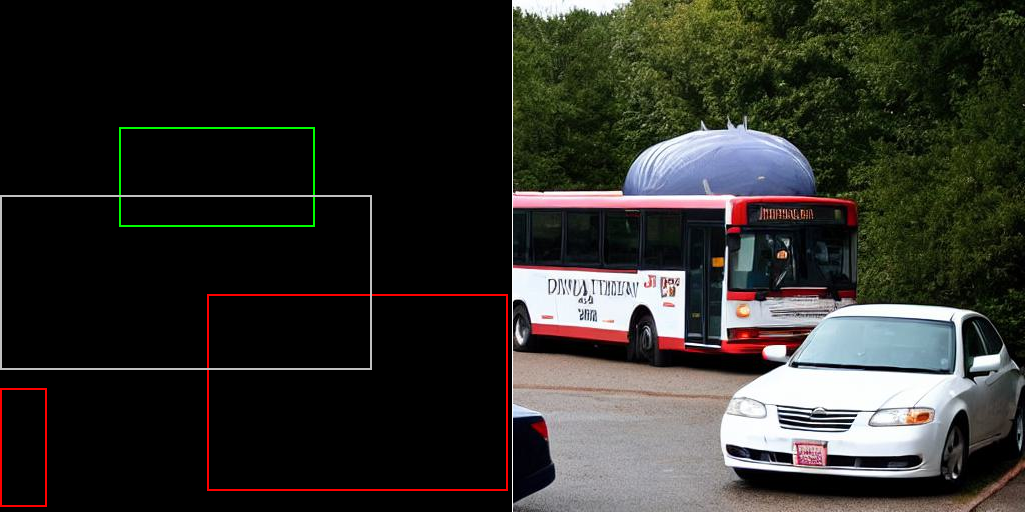}
\subcaption[first caption.]{
\underline{Image Caption:} “There is a red car parked next to an old bus" \\\underline{Conditional Entities:} \textcolor{myred}{car}, \textcolor{mygreen}{boat}, \textcolor{mysilver}{bus}
}\label{fig:2a}
\end{minipage}%
\hspace{0.001\textwidth}
\begin{minipage}[t]{0.327\textwidth}
  \centering
  \captionsetup{font=small} 
\includegraphics[width=1\textwidth]{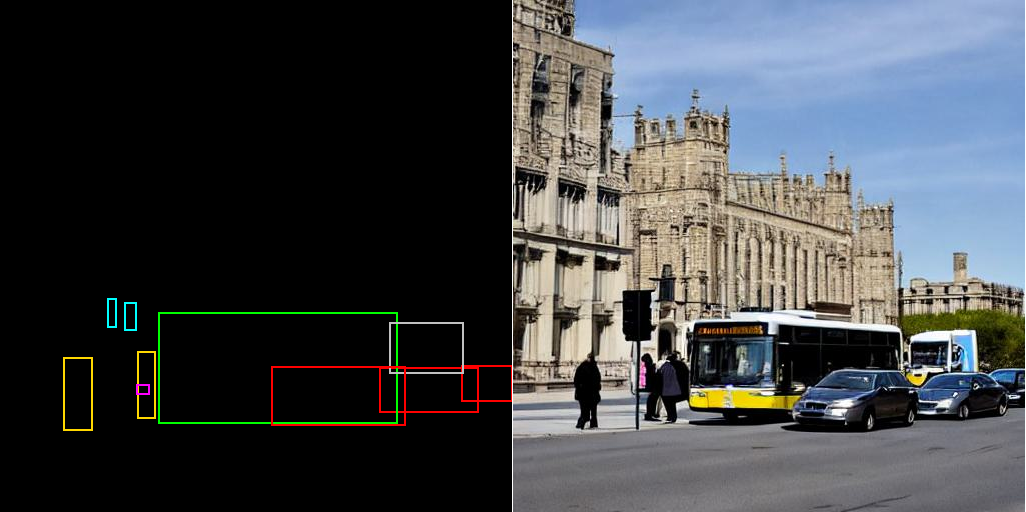}
\subcaption[second caption.]{
\underline{Image Caption:} “An intersection with antique cars and a bus at it" \\\underline{Conditional Entities:} \textcolor{myred}{car}, \textcolor{mygreen}{bus}, \textcolor{mysilver}{truck}, \textcolor{mycyan}{traffic light}, \textcolor{mygold}{person}, \textcolor{mymagenta}{book}
}\label{fig:2b}
\end{minipage}%
\hspace{0.001\textwidth}
\begin{minipage}[t]{0.327\textwidth}
  \centering
   \captionsetup{font=small} 
\includegraphics[width=1\textwidth]{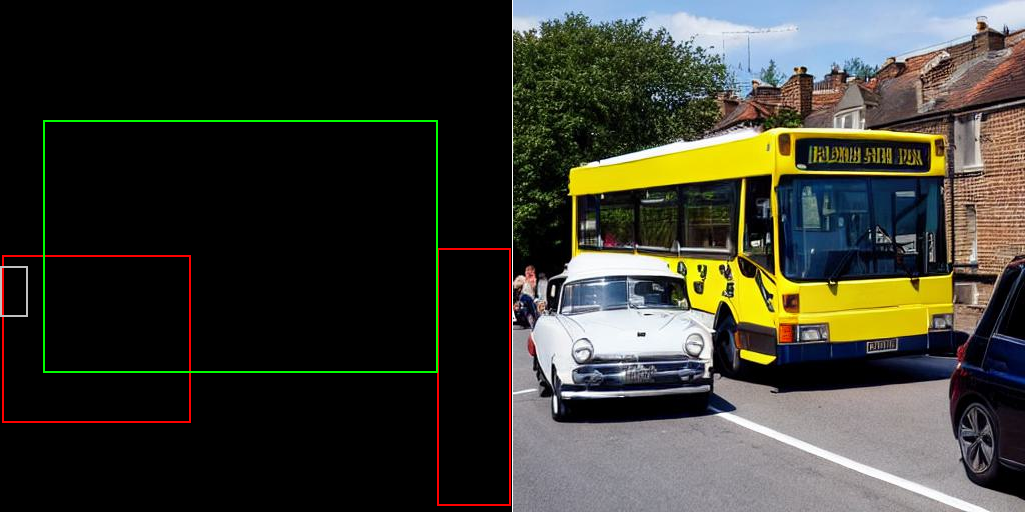}
\subcaption[third caption.]{
\underline{Image Caption:} “a bus and a car on a city street" \\\underline{Conditional Entities:} \textcolor{myred}{car}, \textcolor{mygreen}{bus}, \textcolor{mysilver}{person}
 }\label{fig:2c}
\end{minipage}

\vspace{0.3cm}

\centering
\begin{minipage}[t]{0.327\textwidth}
  \centering
  \captionsetup{font=small} 
\includegraphics[width=1\textwidth]{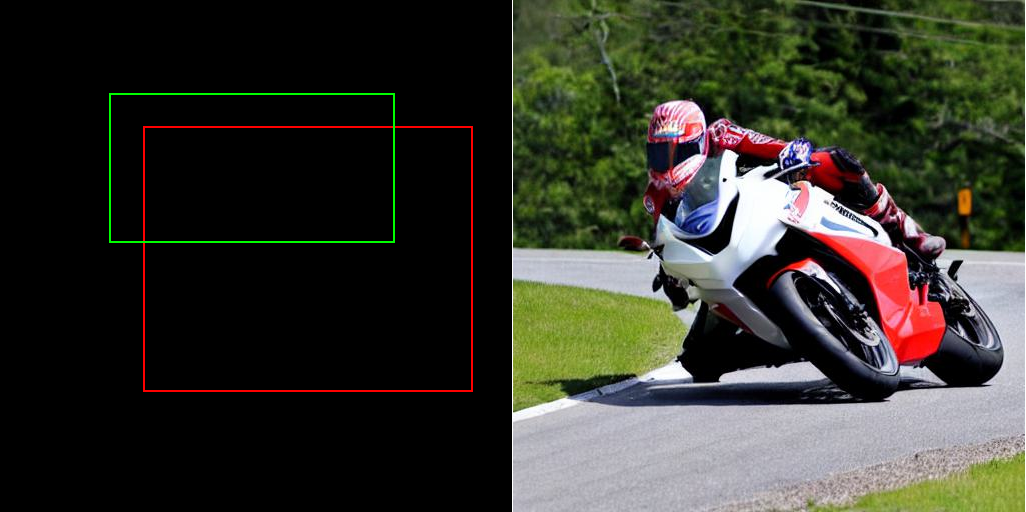}
\subcaption[first caption.]{
\underline{Image Caption:} “A person is leaning low on their motorcycle on the tracks" \\\underline{Conditional Entities:} \textcolor{myred}{motorcycle}, \textcolor{mygreen}{person}
}\label{fig:2a}
\end{minipage}%
\hspace{0.001\textwidth}
\begin{minipage}[t]{0.327\textwidth}
  \centering
  \captionsetup{font=small} 
\includegraphics[width=1\textwidth]{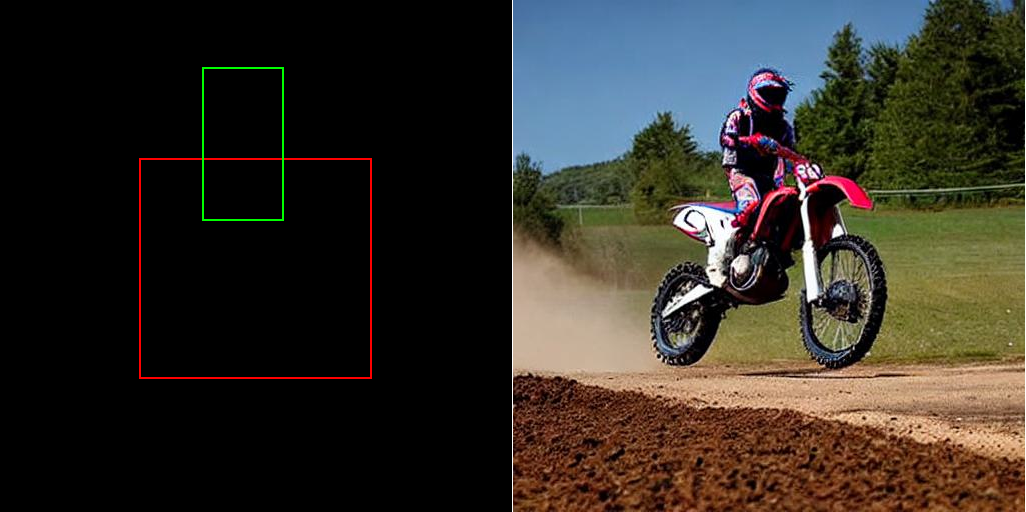}
\subcaption[second caption.]{
\underline{Image Caption:} “A person on a motorcycle in the dirt doing tricks" \\\underline{Conditional Entities:} \textcolor{myred}{motorcycle}, \textcolor{mygreen}{person}
}\label{fig:2b}
\end{minipage}%
\hspace{0.001\textwidth}
\begin{minipage}[t]{0.327\textwidth}
  \centering
   \captionsetup{font=small} 
\includegraphics[width=1\textwidth]{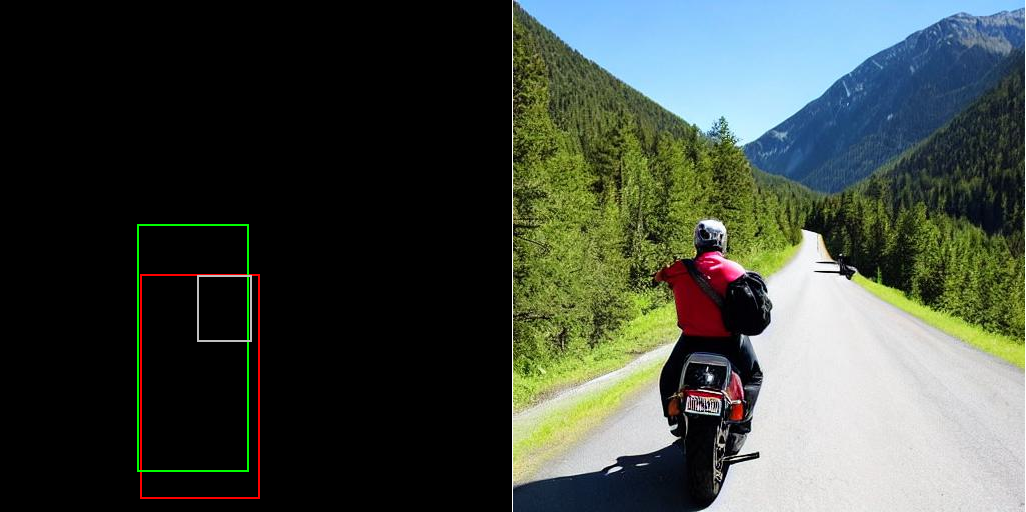}
\subcaption[third caption.]{
\underline{Image Caption:} “A man riding a motorcycle on the side of a road" \\\underline{Conditional Entities:} \textcolor{myred}{motorcycle}, \textcolor{mygreen}{person}, \textcolor{mysilver}{backpack}
}\label{fig:2c}
\end{minipage}

\vspace{0.3cm}

\centering
\begin{minipage}[t]{0.327\textwidth}
  \centering
  \captionsetup{font=small} 
\includegraphics[width=1\textwidth]{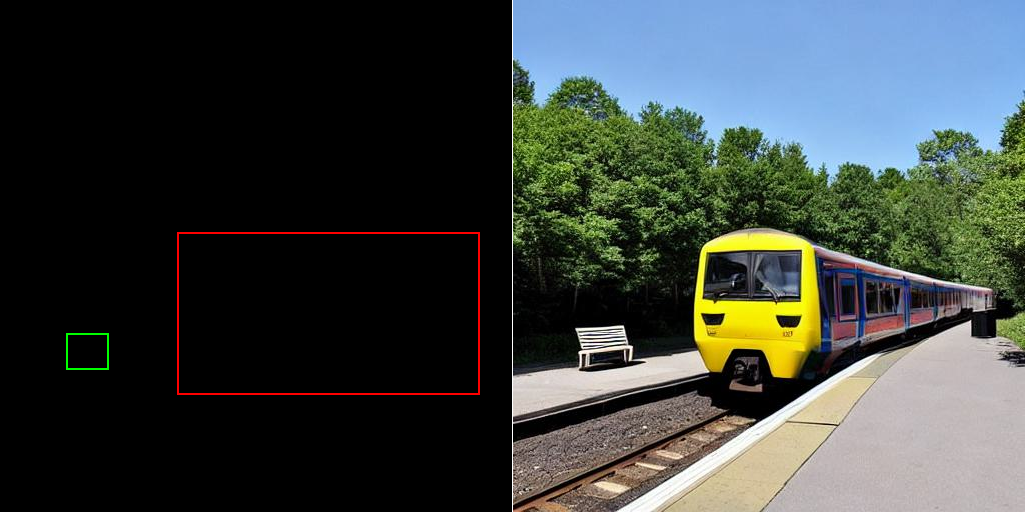}
\subcaption[first caption.]{
\underline{Image Caption:} “Single train parked at a train station on a clear day" \\\underline{Conditional Entities:} \textcolor{myred}{train}, \textcolor{mygreen}{bench}
}\label{fig:2a}
\end{minipage}%
\hspace{0.001\textwidth}
\begin{minipage}[t]{0.327\textwidth}
  \centering
  \captionsetup{font=small} 
\includegraphics[width=1\textwidth]{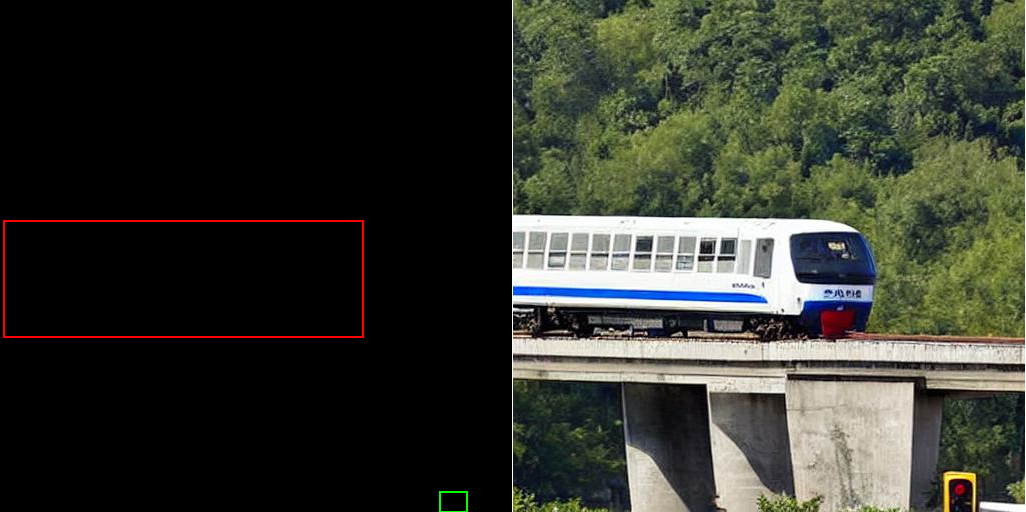}
\subcaption[second caption.]{
\underline{Image Caption:} “A train as it travels down the tracks over a bridge" \\\underline{Conditional Entities:} \textcolor{myred}{train}, \textcolor{mygreen}{traffic light}
}\label{fig:2b}
\end{minipage}%
\hspace{0.001\textwidth}
\begin{minipage}[t]{0.327\textwidth}
  \centering
   \captionsetup{font=small} 
\includegraphics[width=1\textwidth]{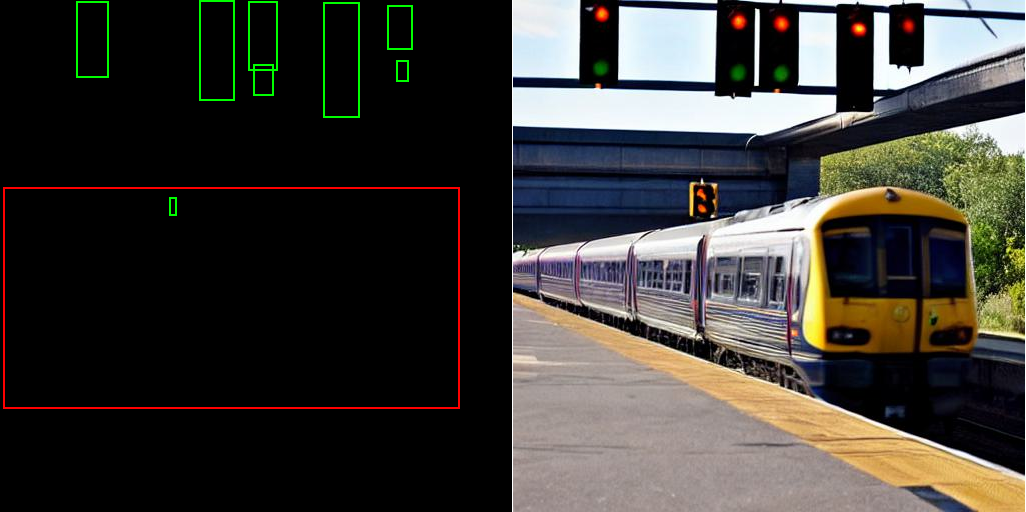}
\subcaption[third caption.]{
\underline{Image Caption:} “A train traveling under a signal lights on top of tracks" \\\underline{Conditional Entities:} \textcolor{myred}{train}, \textcolor{mygreen}{traffic light}
}\label{fig:2c}
\end{minipage}

\vspace{0.3cm}

\centering
\begin{minipage}[t]{0.327\textwidth}
  \centering
  \captionsetup{font=small} 
\includegraphics[width=1\textwidth]{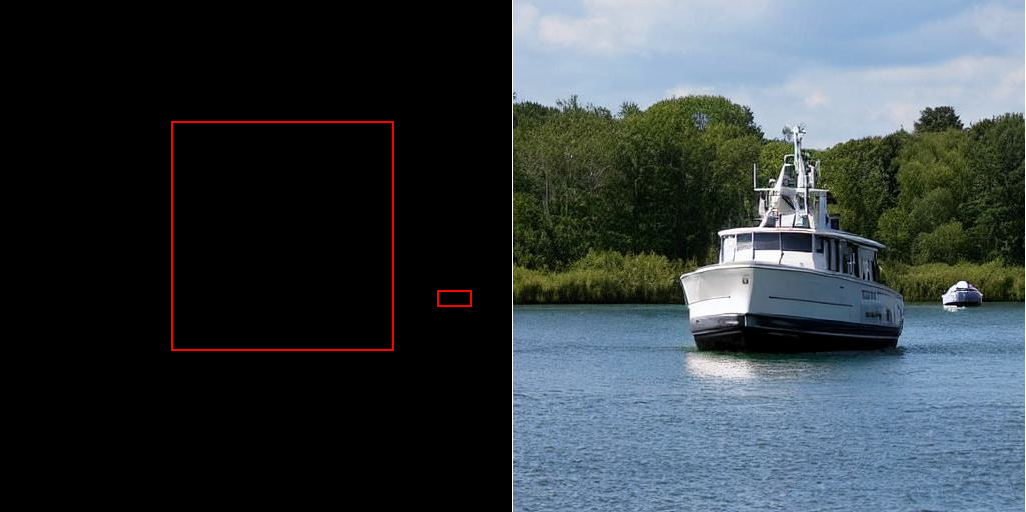}
\subcaption[first caption.]{
\underline{Image Caption:} “a boat is sitting next to a dock" \\\underline{Conditional Entities:} \textcolor{myred}{boat}
}\label{fig:2a}
\end{minipage}%
\hspace{0.001\textwidth}
\begin{minipage}[t]{0.327\textwidth}
  \centering
  \captionsetup{font=small} 
\includegraphics[width=1\textwidth]{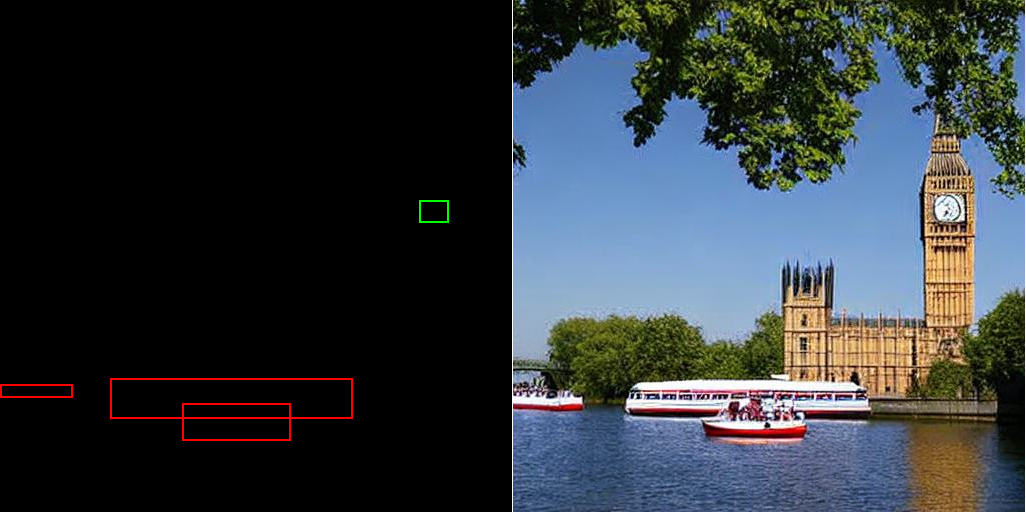}
\subcaption[second caption.]{
 \underline{Image Caption:} “a castle and the big ben clocktower next to a river" \\\underline{Conditional Entities:} \textcolor{myred}{boat}, \textcolor{mygreen}{clock}
}\label{fig:2b}
\end{minipage}%
\hspace{0.001\textwidth}
\begin{minipage}[t]{0.327\textwidth}
  \centering
   \captionsetup{font=small} 
\includegraphics[width=1\textwidth]{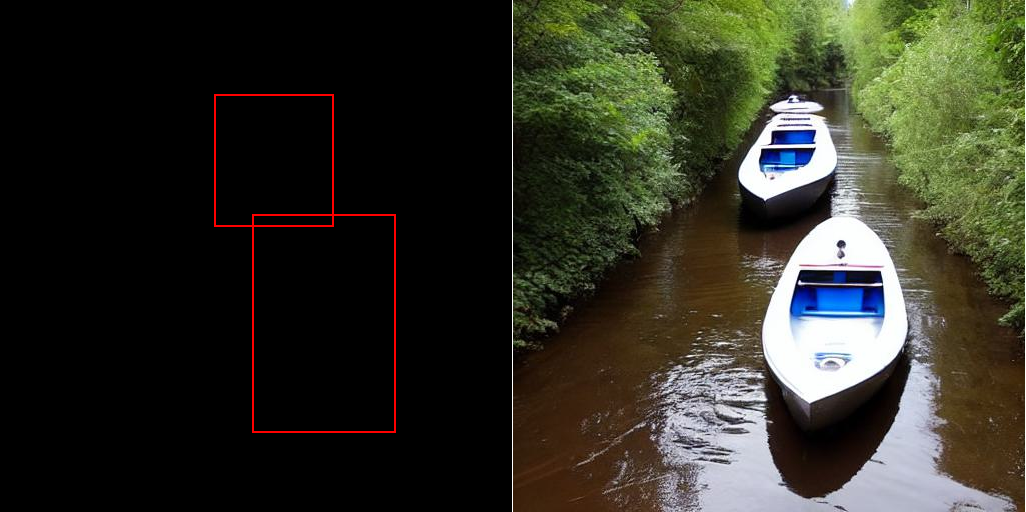}
\subcaption[third caption.]{
\underline{Image Caption:} “Two row boats are laying upside down in the woods" \\\underline{Conditional Entities:} \textcolor{myred}{boat}
}\label{fig:2c}
\end{minipage}

\caption{This figure presents qualitative examples of closed-set configuration. Our model is capable of accurately rendering various classes of vehicles, such as cars, trains, motorcycles, and boats. Images are generated using grounding entities and image captions from the COCO2017 \cite{lin2014microsoft} validation set annotations.}
\label{fig:closed-set-vehicles}
\end{figure*}

\begin{figure*}[htbp]

\centering
\begin{minipage}[t]{0.327\textwidth}
  \centering
  \captionsetup{font=small} 
\includegraphics[width=1\textwidth]{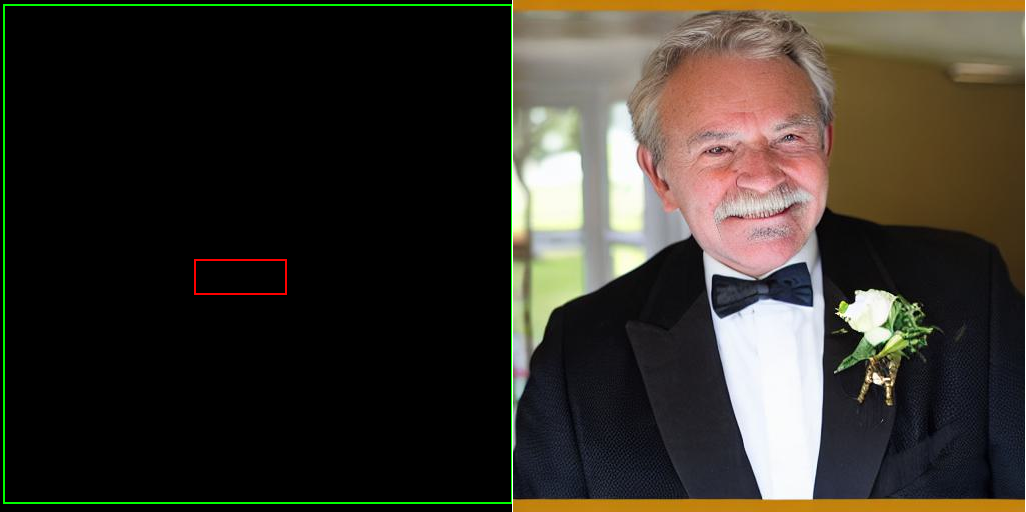}
\subcaption[first caption.]{
\ \underline{Image Caption:} “An older man with a bow tie happily poses for a picture" \\\underline{Conditional Entities:} \textcolor{myred}{tie}, \textcolor{mygreen}{person}
}\label{fig:3a}
\end{minipage}%
\hspace{0.001\textwidth}
\begin{minipage}[t]{0.327\textwidth}
  \centering
  \captionsetup{font=small} 
\includegraphics[width=1\textwidth]{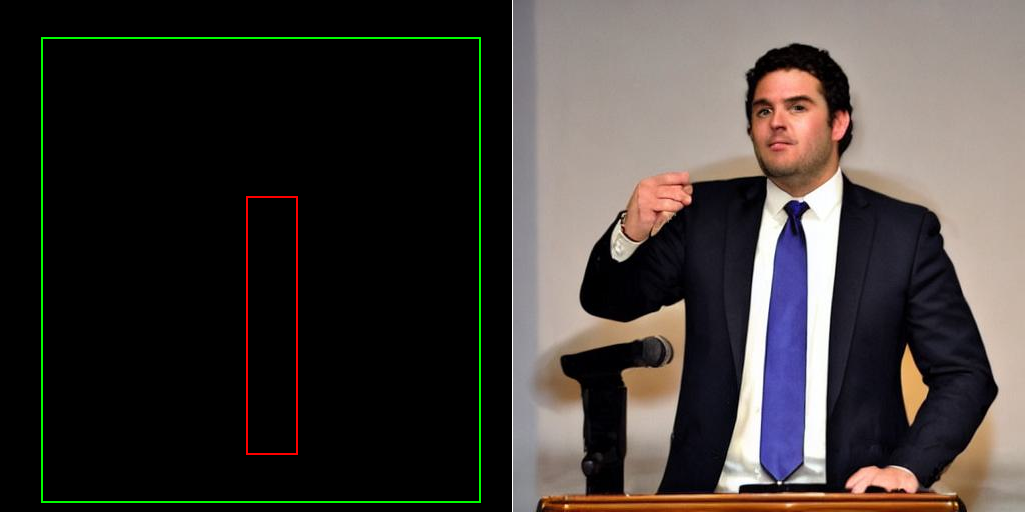}
\subcaption[second caption.]{
\underline{Image Caption:} “A man in a suit is wearing a lit up tie" \\\underline{Conditional Entities:} \textcolor{myred}{tie}, \textcolor{mygreen}{person}
}\label{fig:3b}
\end{minipage}%
\hspace{0.001\textwidth}
\begin{minipage}[t]{0.327\textwidth}
  \centering
   \captionsetup{font=small} 
\includegraphics[width=1\textwidth]{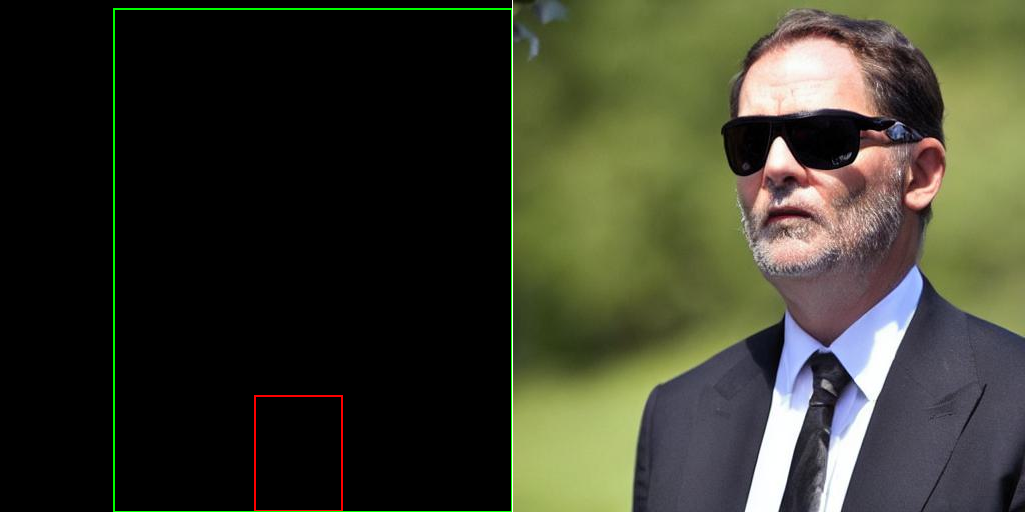}
\subcaption[third caption.]{
\underline{Image Caption:} “A man wearing sunglasses and a black hat" \\\underline{Conditional Entities:} \textcolor{myred}{tie}, \textcolor{mygreen}{person}
 }\label{fig:3c}
\end{minipage}

\vspace{0.3cm}

\centering
\begin{minipage}[t]{0.327\textwidth}
  \centering
  \captionsetup{font=small} 
\includegraphics[width=1\textwidth]{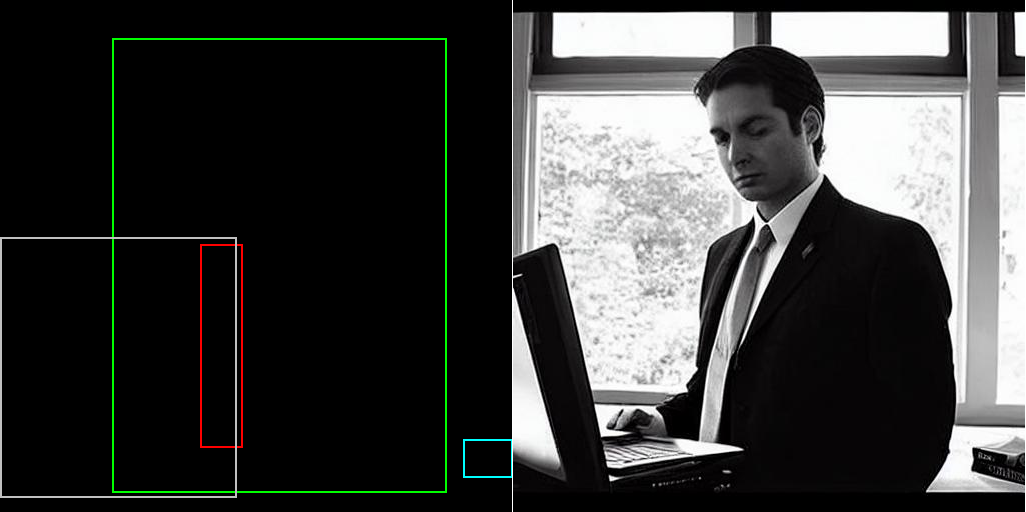}
\subcaption[first caption.]{
\underline{Image Caption:} “a man sitting at a desk in front of a laptop computer" \\\underline{Conditional Entities:} \textcolor{myred}{tie}, \textcolor{mygreen}{person}, \textcolor{mysilver}{laptop}, \textcolor{mycyan}{book}
}\label{fig:3a}
\end{minipage}%
\hspace{0.001\textwidth}
\begin{minipage}[t]{0.327\textwidth}
  \centering
  \captionsetup{font=small} 
\includegraphics[width=1\textwidth]{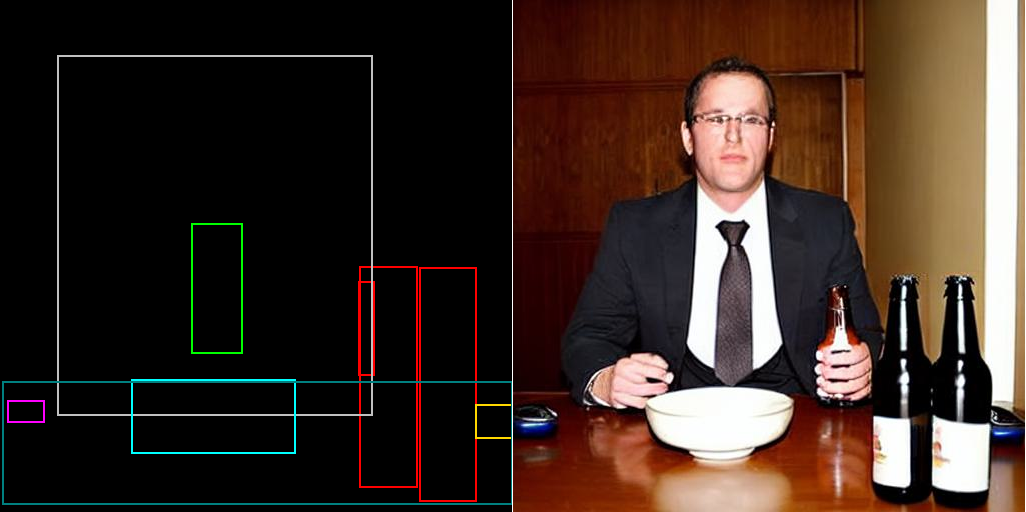}
\subcaption[second caption.]{
\underline{Image Caption:} “am empty bowl and three empty beer bottles in front of a man" \\\underline{Conditional Entities:} \textcolor{myred}{bottle}, \textcolor{mygreen}{tie}, \textcolor{mysilver}{person}, \textcolor{mycyan}{bowl}, \textcolor{mygold}{remote}, \textcolor{mymagenta}{cell phone}, \textcolor{myteal}{dining table}
}\label{fig:3b}
\end{minipage}%
\hspace{0.001\textwidth}
\begin{minipage}[t]{0.327\textwidth}
  \centering
   \captionsetup{font=small} 
\includegraphics[width=1\textwidth]{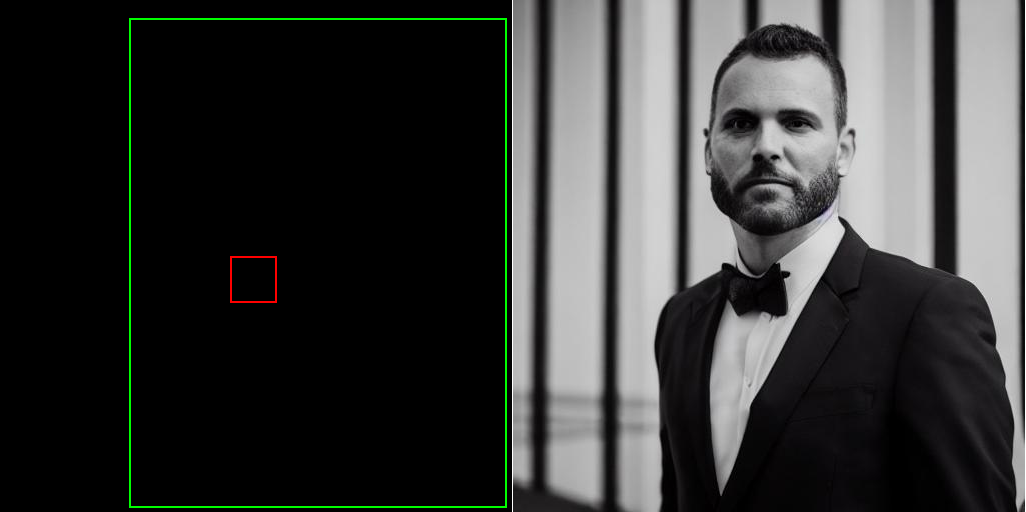}
\subcaption[third caption.]{
\underline{Image Caption:} “a close up of a person wearing a suit and tie" \\\underline{Conditional Entities:} \textcolor{myred}{tie}, \textcolor{mygreen}{person}
}\label{fig:3c}
\end{minipage}

\vspace{0.3cm}

\centering
\begin{minipage}[t]{0.327\textwidth}
  \centering
  \captionsetup{font=small} 
\includegraphics[width=1\textwidth]{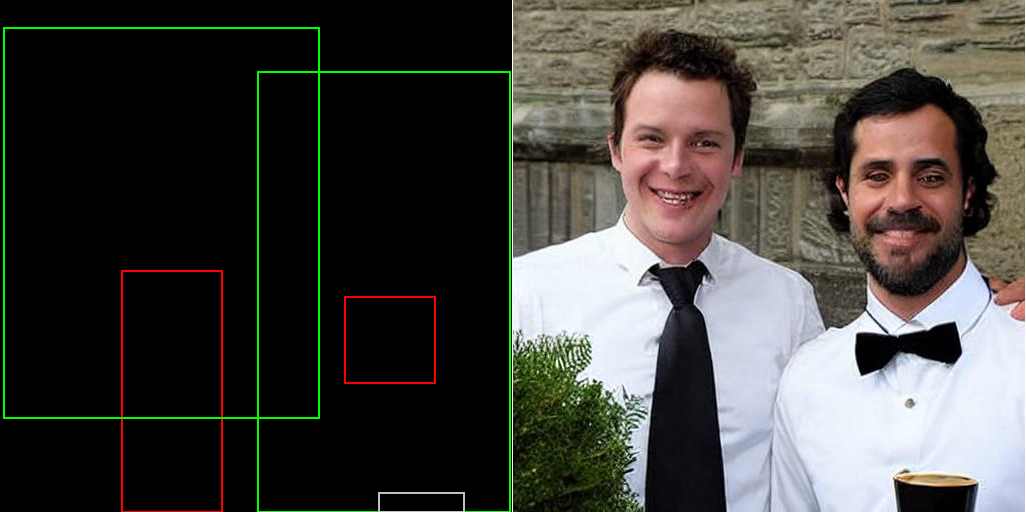}
\subcaption[first caption.]{
\underline{Image Caption:} “Two men smiling and taking a picture together"  \\\underline{Conditional Entities:} \textcolor{myred}{tie}, \textcolor{mygreen}{person}, \textcolor{mysilver}{cup}
}\label{fig:3a}
\end{minipage}%
\hspace{0.001\textwidth}
\begin{minipage}[t]{0.327\textwidth}
  \centering
  \captionsetup{font=small} 
\includegraphics[width=1\textwidth]{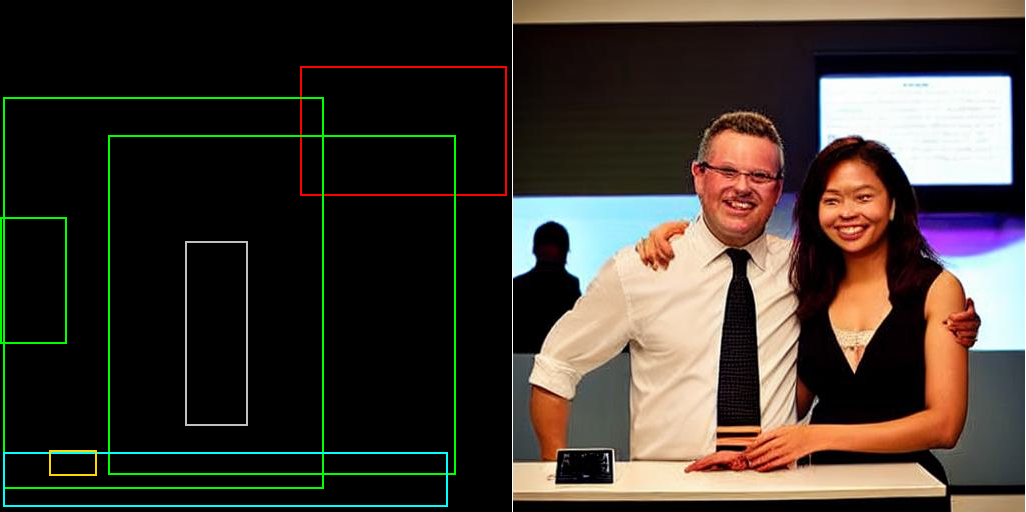}
\subcaption[second caption.]{
\underline{Image Caption:} “A man and a women posing next to one another in front of a table" \\\underline{Conditional Entities:} \textcolor{myred}{tv}, \textcolor{mygreen}{person}, \textcolor{mysilver}{tie}, \textcolor{mycyan}{dining table}, \textcolor{mygold}{cell phone}
}\label{fig:3b}
\end{minipage}%
\hspace{0.001\textwidth}
\begin{minipage}[t]{0.327\textwidth}
  \centering
   \captionsetup{font=small} 
\includegraphics[width=1\textwidth]{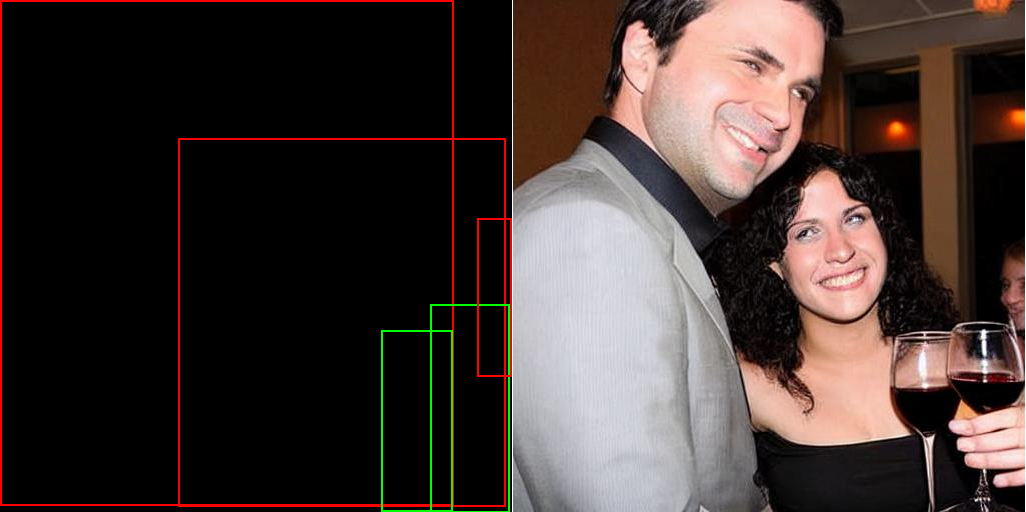}
\subcaption[third caption.]{
\underline{Image Caption:} “A couple of women sitting next to each other" \\\underline{Conditional Entities:} \textcolor{myred}{person}, \textcolor{mygreen}{wine glass}
}\label{fig:3c}
\end{minipage}

\vspace{0.3cm}

\centering
\begin{minipage}[t]{0.327\textwidth}
  \centering
  \captionsetup{font=small} 
\includegraphics[width=1\textwidth]{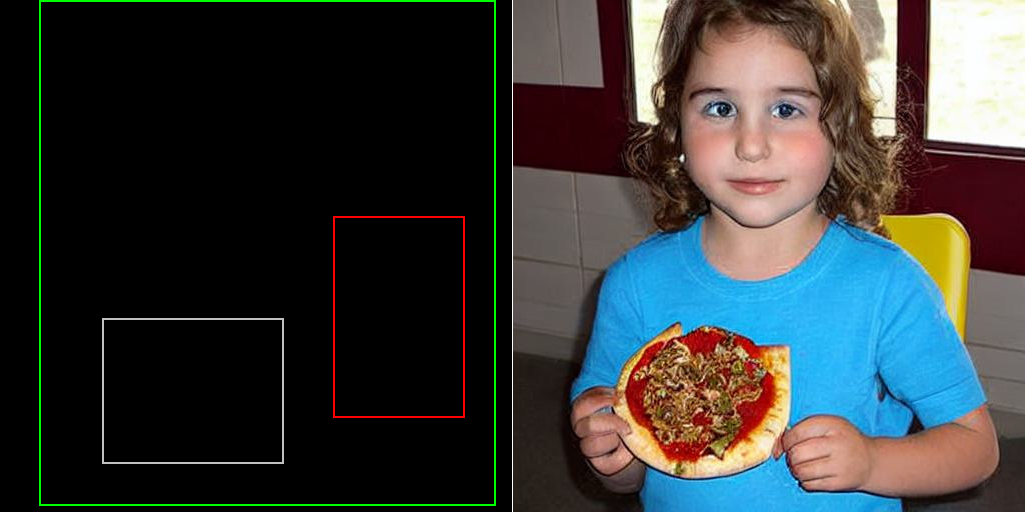}
\subcaption[first caption.]{
\underline{Image Caption:} “A young girl eating a slice of pizza" \\\underline{Conditional Entities:} \textcolor{myred}{chair}, \textcolor{mygreen}{person}, \textcolor{mysilver}{pizza}
}\label{fig:3a}
\end{minipage}%
\hspace{0.001\textwidth}
\begin{minipage}[t]{0.327\textwidth}
  \centering
  \captionsetup{font=small} 
\includegraphics[width=1\textwidth]{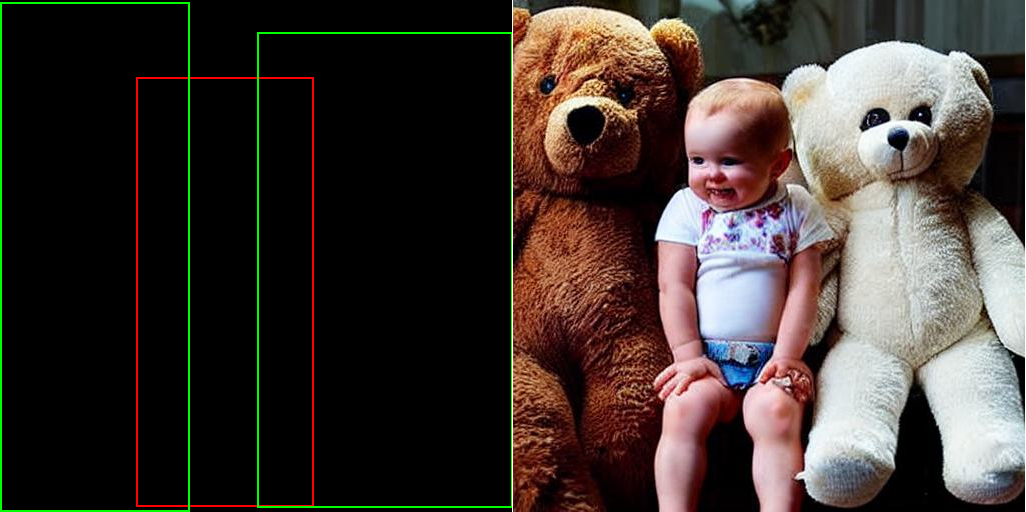}
\subcaption[second caption.]{
\underline{Image Caption:} “A baby sitting in between to large stuffed animals" \\\underline{Conditional Entities:} \textcolor{myred}{person}, \textcolor{mygreen}{teddy bear}
}\label{fig:3b}
\end{minipage}%
\hspace{0.001\textwidth}
\begin{minipage}[t]{0.327\textwidth}
  \centering
   \captionsetup{font=small} 
\includegraphics[width=1\textwidth]{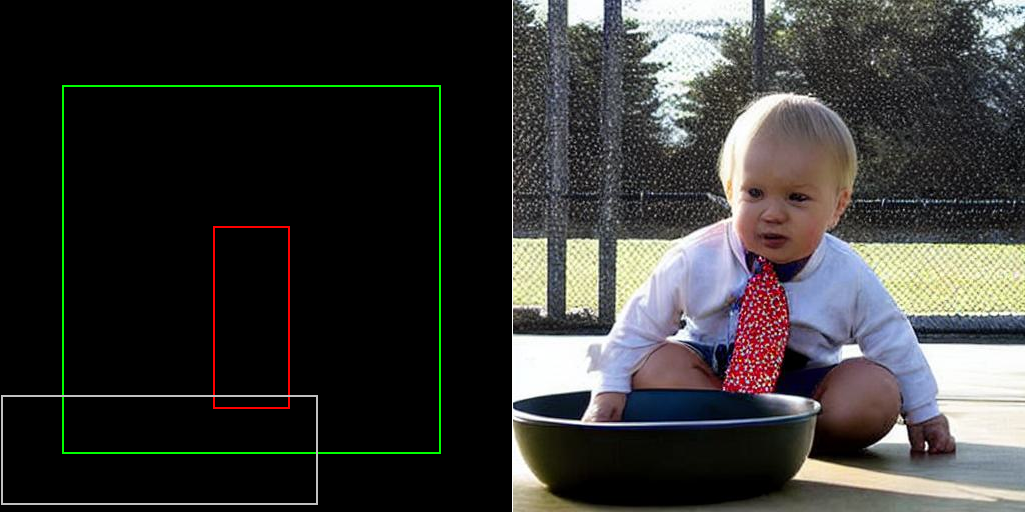}
\subcaption[third caption.]{
\underline{Image Caption:} “A young child squats down in front of a bowl" \\\underline{Conditional Entities:} \textcolor{myred}{tie}, \textcolor{mygreen}{person}, \textcolor{mysilver}{bowl}
}\label{fig:3c}
\end{minipage}
\caption{This figure presents qualitative examples of closed-set configuration. The portraits generated by ObjectDiffusion exhibit realistic facial expressions. Images are generated using grounding entities and image captions from the COCO2017 \cite{lin2014microsoft} validation set annotations.}
\label{fig:closed-set-people}
\end{figure*}

\begin{figure*}[htbp]

\centering
\begin{minipage}[t]{0.327\textwidth}
  \centering
  \captionsetup{font=small} 
\includegraphics[width=1\textwidth]{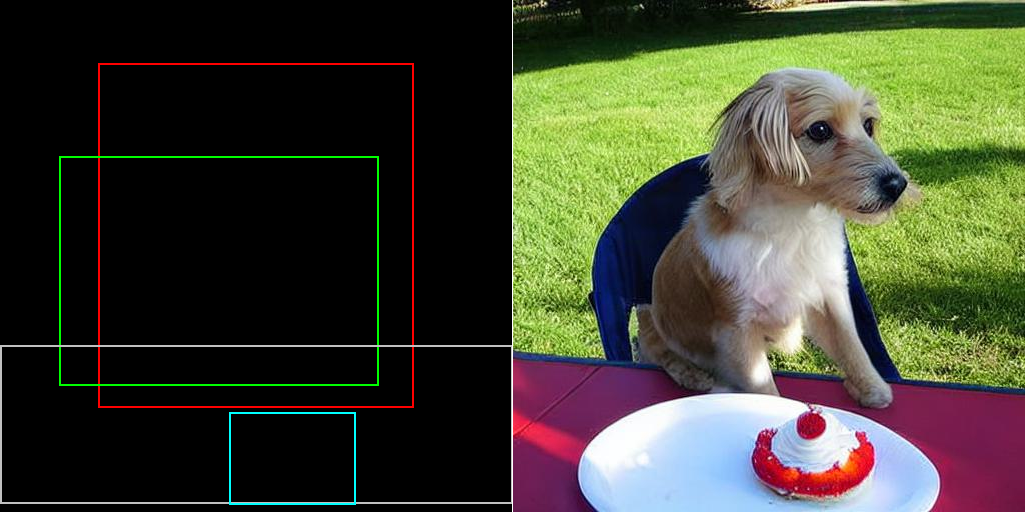}
\subcaption[first caption.]{
\underline{Image Caption:} “A very cute looking small dog by some food" \\\underline{Conditional Entities:} \textcolor{myred}{dog}, \textcolor{mygreen}{chair}, \textcolor{mysilver}{dining table}, \textcolor{mycyan}{cake}
}\label{fig:4a}
\end{minipage}%
\hspace{0.001\textwidth}
\begin{minipage}[t]{0.327\textwidth}
  \centering
  \captionsetup{font=small} 
\includegraphics[width=1\textwidth]{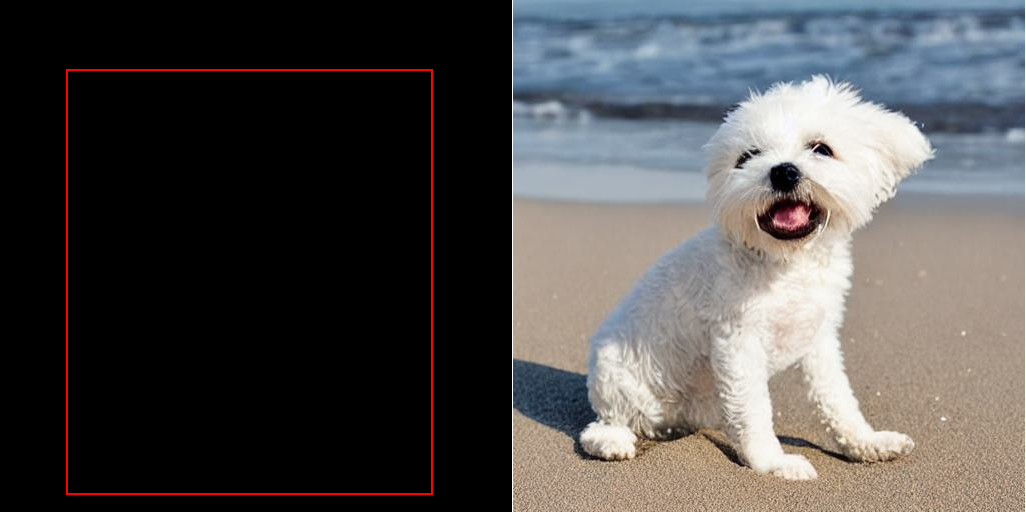}
\subcaption[second caption.]{
 \underline{Image Caption:} “A white dog is on a sandy beach while the sea foam washes ashore behind it" \\\underline{Conditional Entities:} \textcolor{myred}{dog}
}\label{fig:4b}
\end{minipage}%
\hspace{0.001\textwidth}
\begin{minipage}[t]{0.327\textwidth}
  \centering
   \captionsetup{font=small} 
\includegraphics[width=1\textwidth]{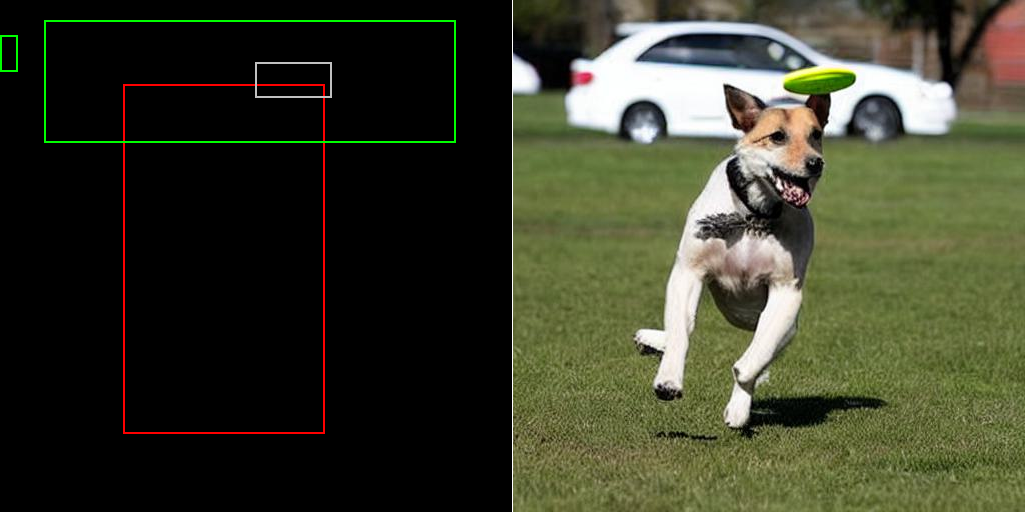}
\subcaption[third caption.]{
\underline{Image Caption:} “a dog jumping to catch a frisbee in a yard" \\\underline{Conditional Entities:} \textcolor{myred}{dog}, \textcolor{mygreen}{car}, \textcolor{mysilver}{frisbee}
 }\label{fig:4c}
\end{minipage}

\vspace{0.3cm}

\centering
\begin{minipage}[t]{0.327\textwidth}
  \centering
  \captionsetup{font=small} 
\includegraphics[width=1\textwidth]{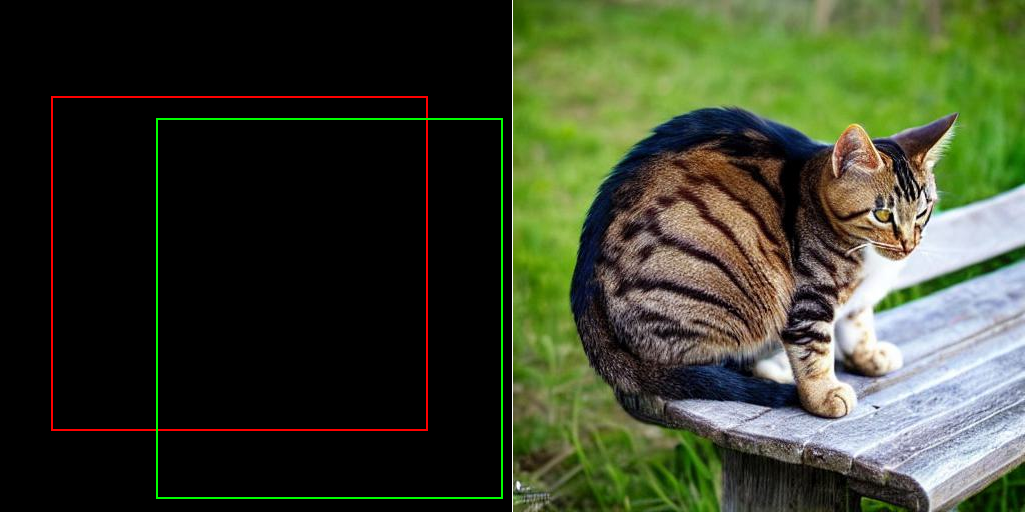}
\subcaption[first caption.]{
\underline{Image Caption:} “A cat sitting on top of a bench in a field" \\\underline{Conditional Entities:} \textcolor{myred}{cat}, \textcolor{mygreen}{bench}
}\label{fig:4a}
\end{minipage}%
\hspace{0.001\textwidth}
\begin{minipage}[t]{0.327\textwidth}
  \centering
  \captionsetup{font=small} 
\includegraphics[width=1\textwidth]{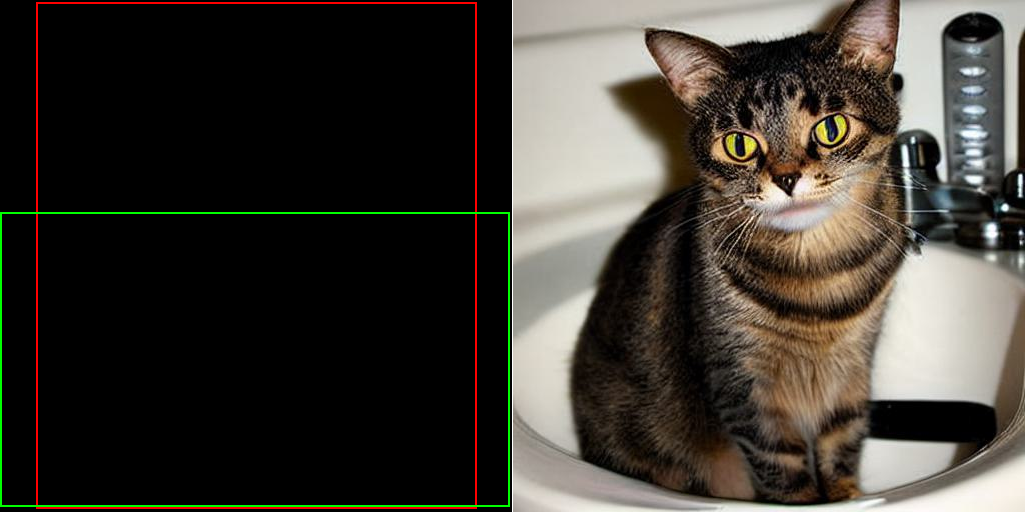}
\subcaption[second caption.]{
\underline{Image Caption:} “A cat sitting in a bathroom sink under a  faucet" \\\underline{Conditional Entities:} \textcolor{myred}{cat}, \textcolor{mygreen}{sink}
}\label{fig:4b}
\end{minipage}%
\hspace{0.001\textwidth}
\begin{minipage}[t]{0.327\textwidth}
  \centering
   \captionsetup{font=small} 
\includegraphics[width=1\textwidth]{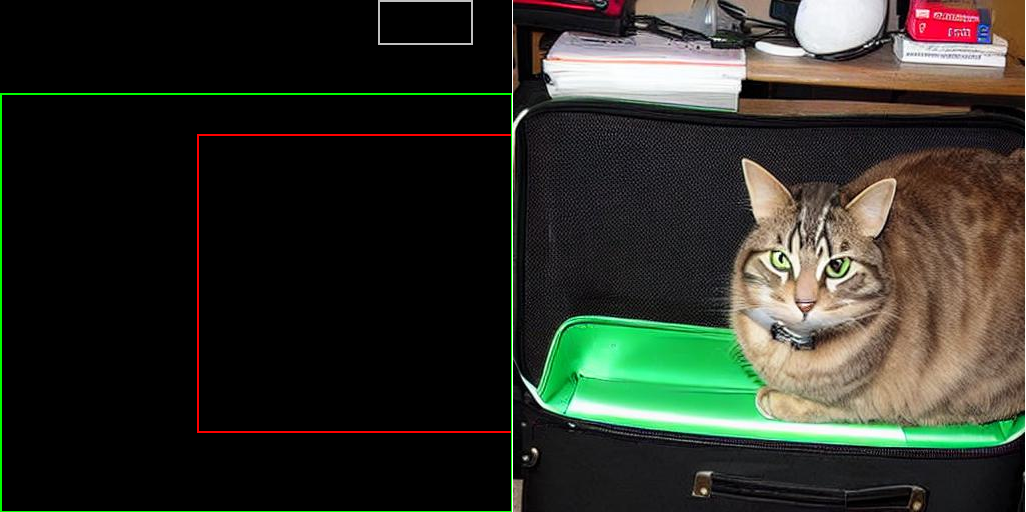}
\subcaption[third caption.]{
 \underline{Image Caption:} “A cat laying on clothes that are in a suitcase"  \\\underline{Conditional Entities:} \textcolor{myred}{cat}, \textcolor{mygreen}{suitcase}, \textcolor{mysilver}{book}
}\label{fig:4c}
\end{minipage}

\vspace{0.3cm}

\centering
\begin{minipage}[t]{0.327\textwidth}
  \centering
  \captionsetup{font=small} 
\includegraphics[width=1\textwidth]{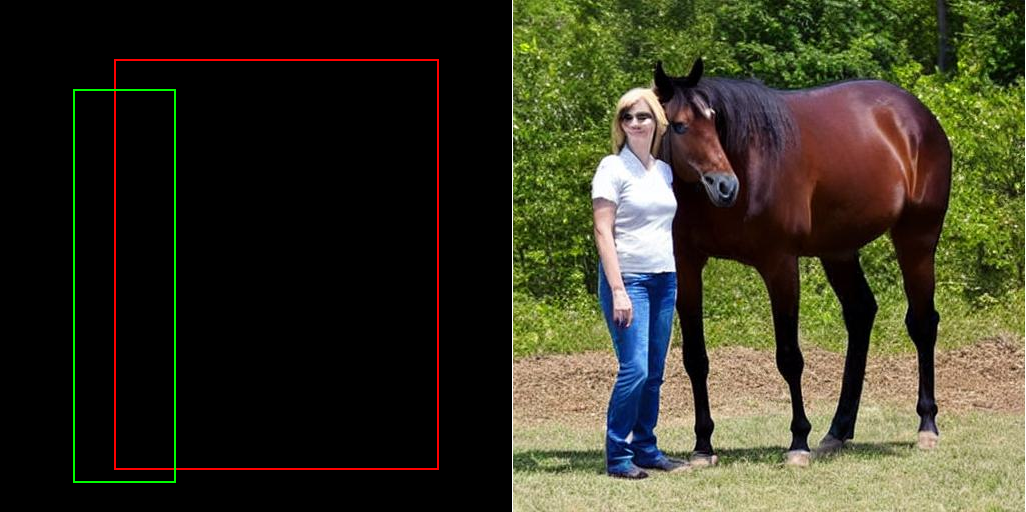}
\subcaption[first caption.]{
\underline{Image Caption:} “A woman in tight blue jeans standing next to a brown horse" \\\underline{Conditional Entities:} \textcolor{myred}{horse}, \textcolor{mygreen}{person}
}\label{fig:4a}
\end{minipage}%
\hspace{0.001\textwidth}
\begin{minipage}[t]{0.327\textwidth}
  \centering
  \captionsetup{font=small} 
\includegraphics[width=1\textwidth]{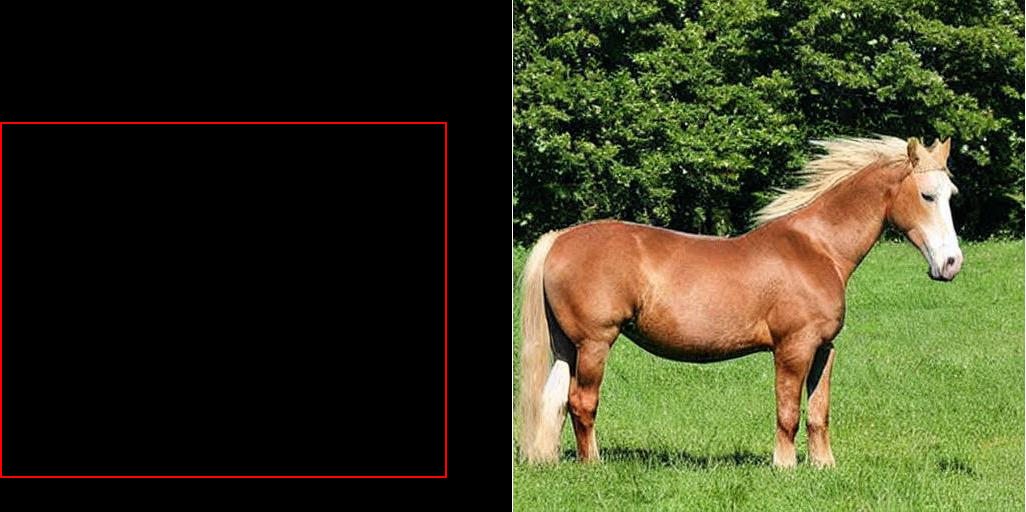}
\subcaption[second caption.]{
 \underline{Image Caption:} “A brown horse with blonde hair standing in an open field" \\\underline{Conditional Entities:} \textcolor{myred}{horse}
}\label{fig:4b}
\end{minipage}%
\hspace{0.001\textwidth}
\begin{minipage}[t]{0.327\textwidth}
  \centering
   \captionsetup{font=small} 
\includegraphics[width=1\textwidth]{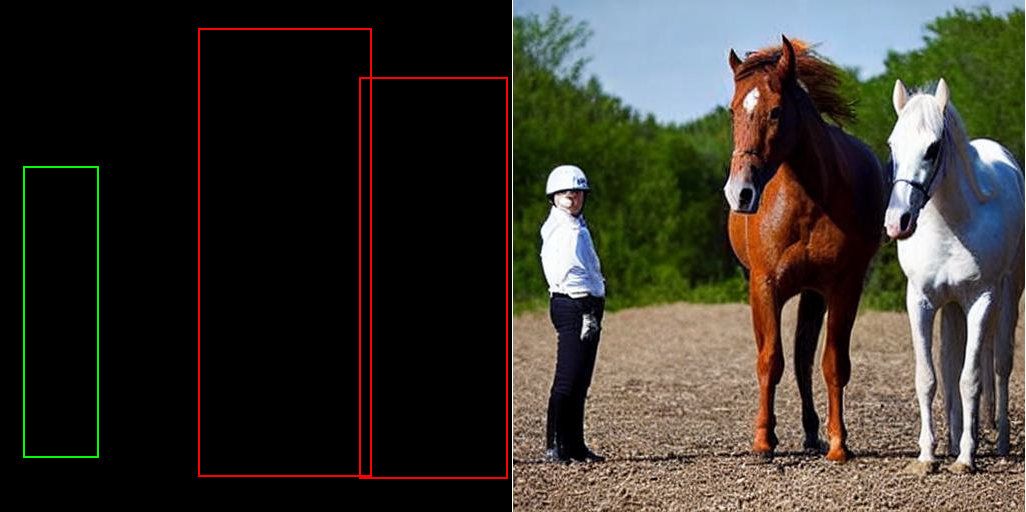}
\subcaption[third caption.]{
\underline{Image Caption:} “a couple of horses pull a contraption" \\\underline{Conditional Entities:} \textcolor{myred}{horse}, \textcolor{mygreen}{person}
}\label{fig:4c}
\end{minipage}

\vspace{0.3cm}

\centering
\begin{minipage}[t]{0.327\textwidth}
  \centering
  \captionsetup{font=small} 
\includegraphics[width=1\textwidth]{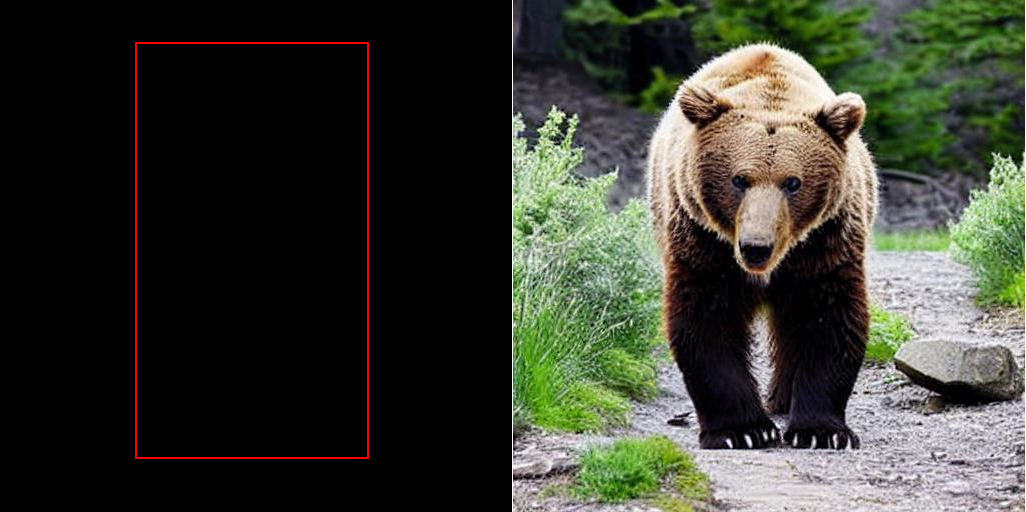}
\subcaption[first caption.]{
\underline{Image Caption:} “two different bears fight with each other behind a log" \\\underline{Conditional Entities:} \textcolor{myred}{bear}
}\label{fig:4a}
\end{minipage}%
\hspace{0.001\textwidth}
\begin{minipage}[t]{0.327\textwidth}
  \centering
  \captionsetup{font=small} 
\includegraphics[width=1\textwidth]{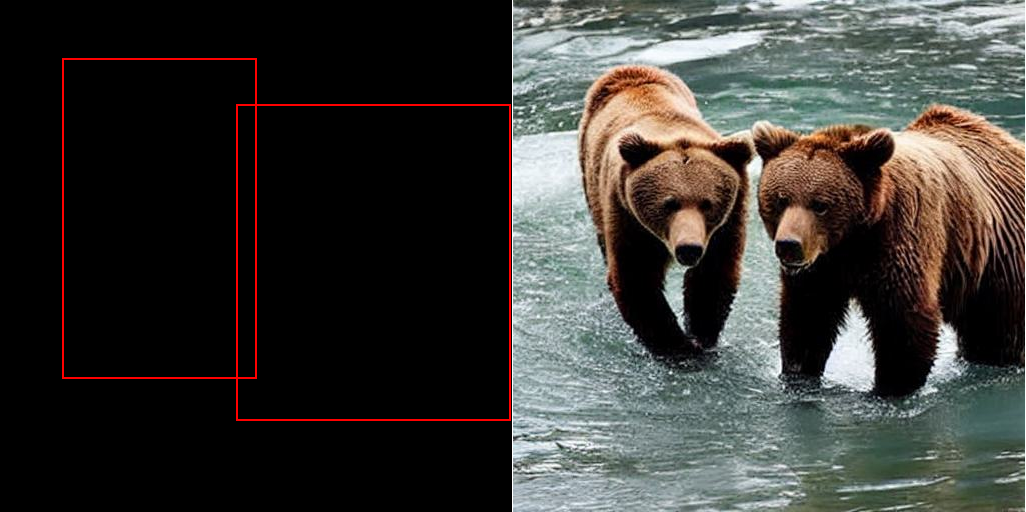}
\subcaption[second caption.]{
 \underline{Image Caption:} “Two brown bears swimming together in the water" \\\underline{Conditional Entities:} \textcolor{myred}{bear}
}\label{fig:4b}
\end{minipage}%
\hspace{0.001\textwidth}
\begin{minipage}[t]{0.327\textwidth}
  \centering
   \captionsetup{font=small} 
\includegraphics[width=1\textwidth]{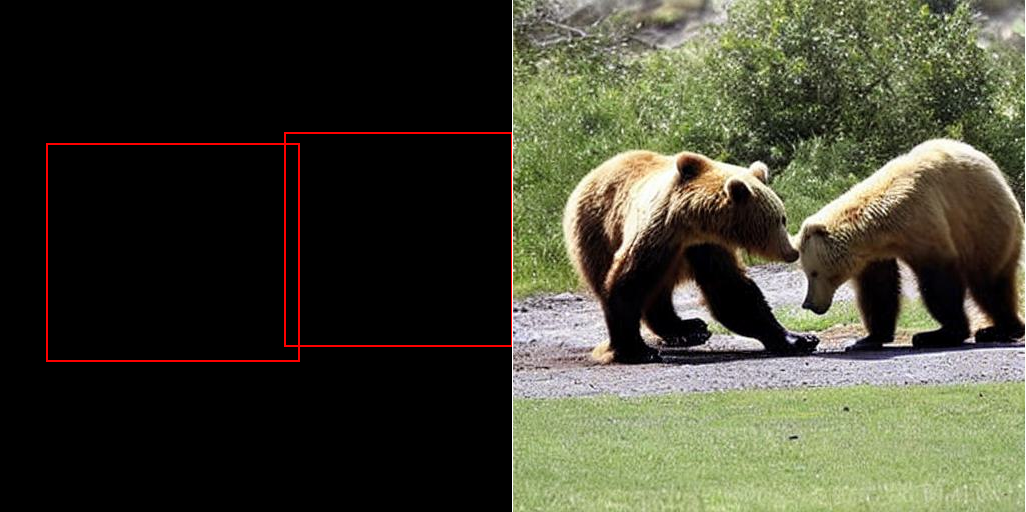}
\subcaption[third caption.]{
\underline{Image Caption:} “two different bears fight with each other behind a log" \\\underline{Conditional Entities:} \textcolor{myred}{bear}
}\label{fig:4c}
\end{minipage}

\caption{This figure represents qualitative examples of closed-set configuration. Our model demonstrates proficiency in generating high-resolution visuals of animals in their native habitat, achieving a very high AP precision score. Images are generated using grounding entities and image captions from COCO2017 \cite{lin2014microsoft} validation set annotations.}
\label{fig:closed-set-animals}
\end{figure*}

\begin{figure*}[htbp]

\centering
\begin{minipage}[t]{0.327\textwidth}
  \centering
  \captionsetup{font=small} 
\includegraphics[width=1\textwidth]{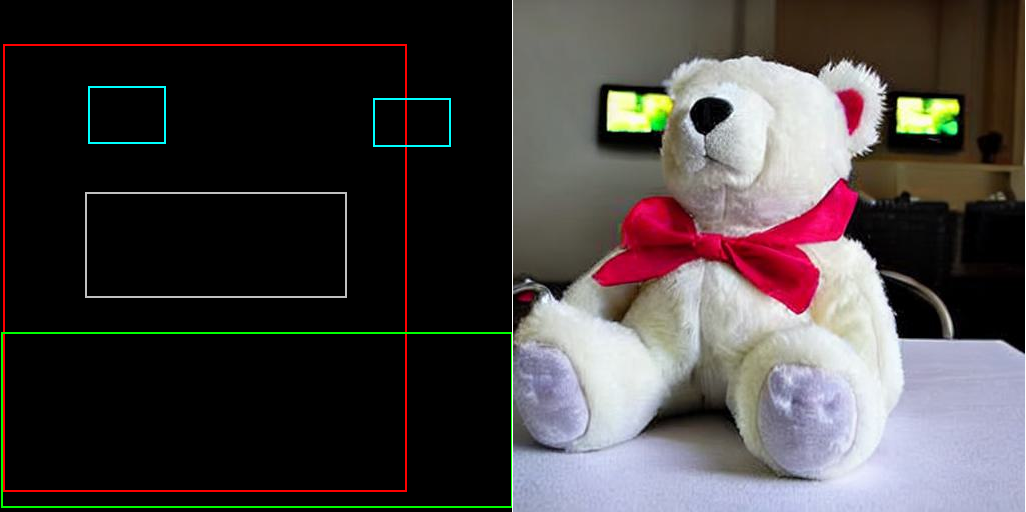}
\subcaption[first caption.]{
\underline{Image Caption:} “A carved bear that has a ribbon around the neck" \\\underline{Conditional Entities:} \textcolor{myred}{teddy bear}, \textcolor{mygreen}{dining table}, \textcolor{mysilver}{tie}, \textcolor{mycyan}{tv}}\label{fig:5a}
\end{minipage}%
\hspace{0.001\textwidth}
\begin{minipage}[t]{0.327\textwidth}
  \centering
  \captionsetup{font=small} 
\includegraphics[width=1\textwidth]{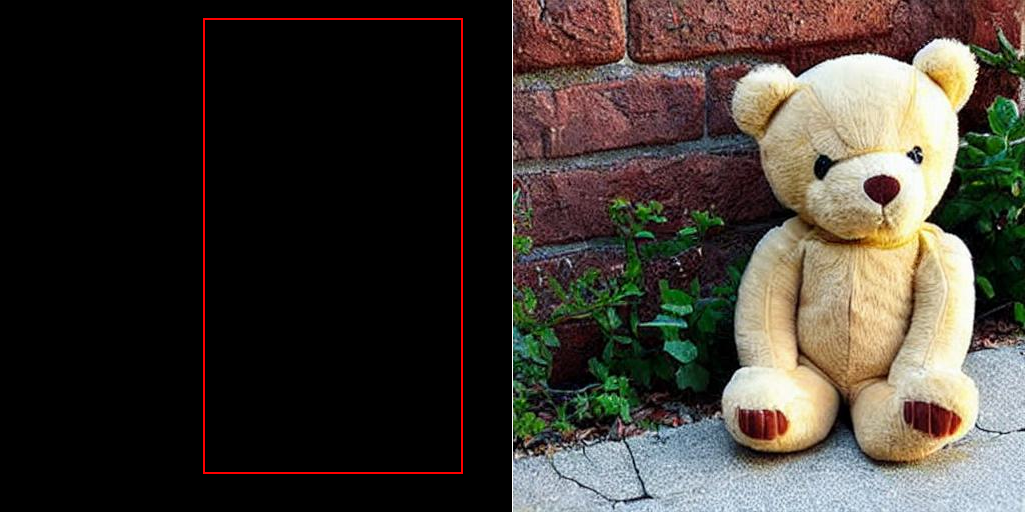}
\subcaption[second caption.]{
\underline{Image Caption:} “The stuffed teddy bear is sitting near the wall" \\\underline{Conditional Entities:} \textcolor{myred}{teddy bear}
}\label{fig:5b}
\end{minipage}%
\hspace{0.001\textwidth}
\begin{minipage}[t]{0.327\textwidth}
  \centering
   \captionsetup{font=small} 
\includegraphics[width=1\textwidth]{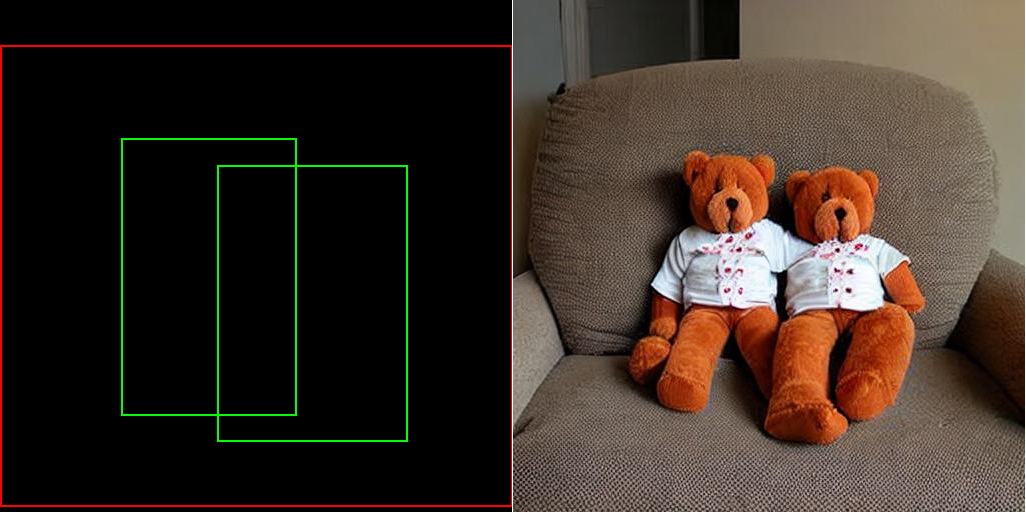}
\subcaption[third caption.]{
 \underline{Image Caption:} “teddy bears dressed up in clothing sitting on a loveseat together" \\\underline{Conditional Entities:} \textcolor{myred}{couch}, \textcolor{mygreen}{teddy bear}}\label{fig:5c}
\end{minipage}

\vspace{0.3cm}

\centering
\begin{minipage}[t]{0.327\textwidth}
  \centering
  \captionsetup{font=small} 
\includegraphics[width=1\textwidth]{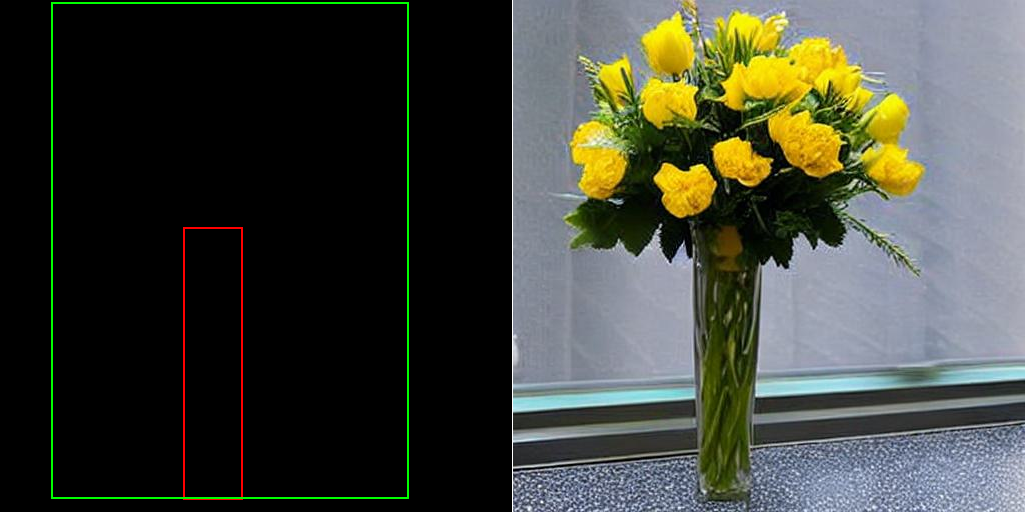}
\subcaption[first caption.]{
 \underline{Image Caption:} “A vase filled  with yellow flowers next to a window" \\\underline{Conditional Entities:} \textcolor{myred}{vase}, \textcolor{mygreen}{potted plant}
}\label{fig:5a}
\end{minipage}%
\hspace{0.001\textwidth}
\begin{minipage}[t]{0.327\textwidth}
  \centering
  \captionsetup{font=small} 
\includegraphics[width=1\textwidth]{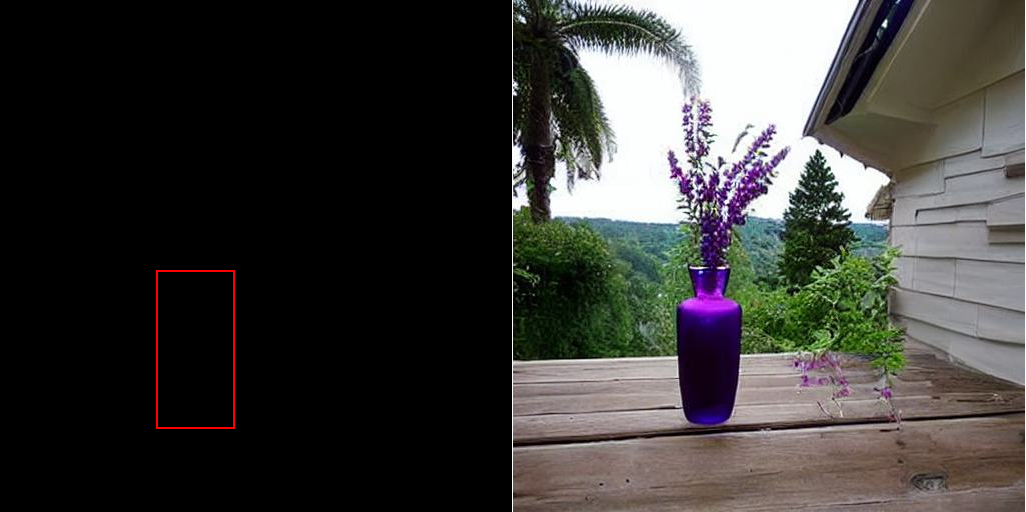}
\subcaption[second caption.]{
\underline{Image Caption:} “there is a small glass vase that has purple flowers in it" \\\underline{Conditional Entities:} \textcolor{myred}{vase}
}\label{fig:5b}
\end{minipage}%
\hspace{0.001\textwidth}
\begin{minipage}[t]{0.327\textwidth}
  \centering
   \captionsetup{font=small} 
\includegraphics[width=1\textwidth]{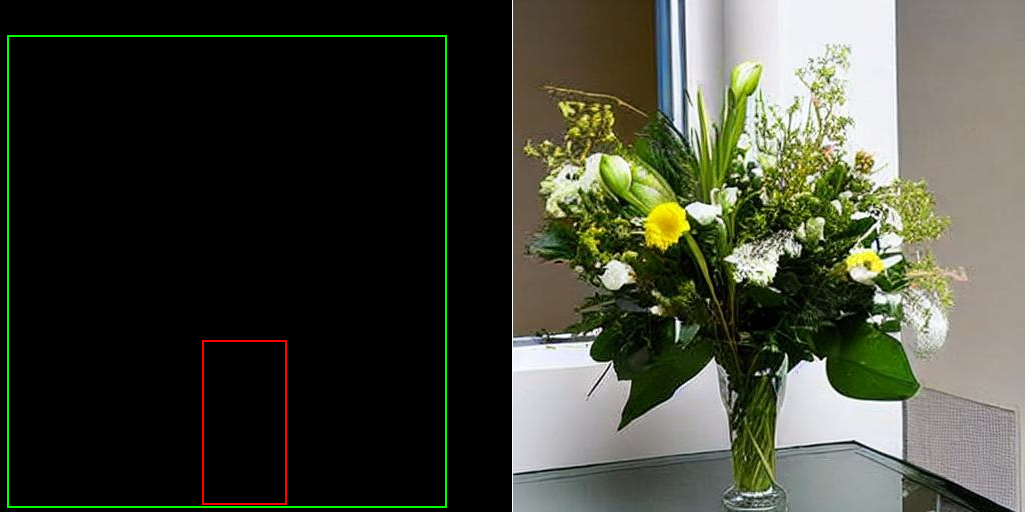}
\subcaption[third caption.]{
\underline{Image Caption:} “a green vase with some flowers and greenery" \\\underline{Conditional Entities:} \textcolor{myred}{vase}, \textcolor{mygreen}{potted plant}
}\label{fig:5c}
\end{minipage}

\vspace{0.3cm}

\centering
\begin{minipage}[t]{0.327\textwidth}
  \centering
  \captionsetup{font=small} 
\includegraphics[width=1\textwidth]{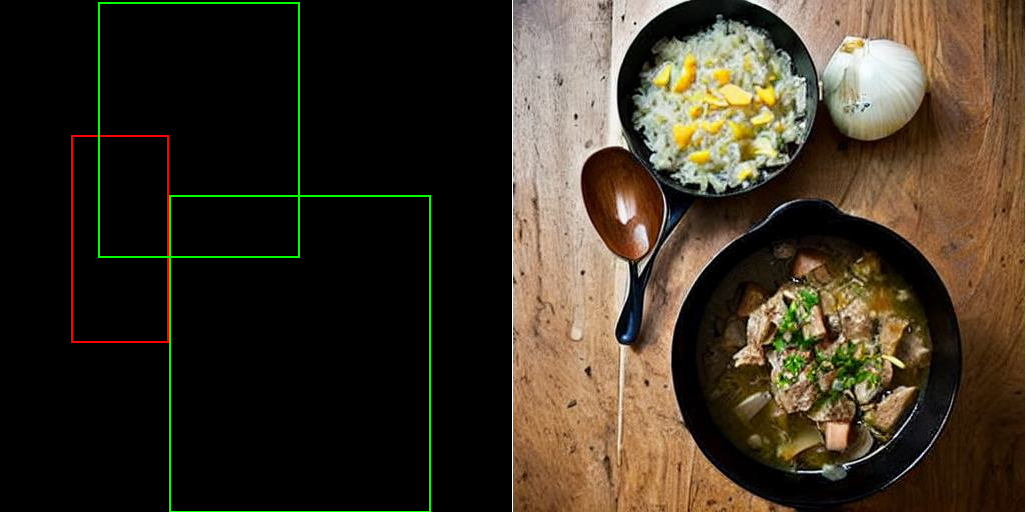}
\subcaption[first caption.]{
 \underline{Image Caption:} “A pan filled with onions sitting next to a pan of stew" \\\underline{Conditional Entities:} \textcolor{myred}{spoon}, \textcolor{mygreen}{bowl}
}\label{fig:5a}
\end{minipage}%
\hspace{0.001\textwidth}
\begin{minipage}[t]{0.327\textwidth}
  \centering
  \captionsetup{font=small} 
\includegraphics[width=1\textwidth]{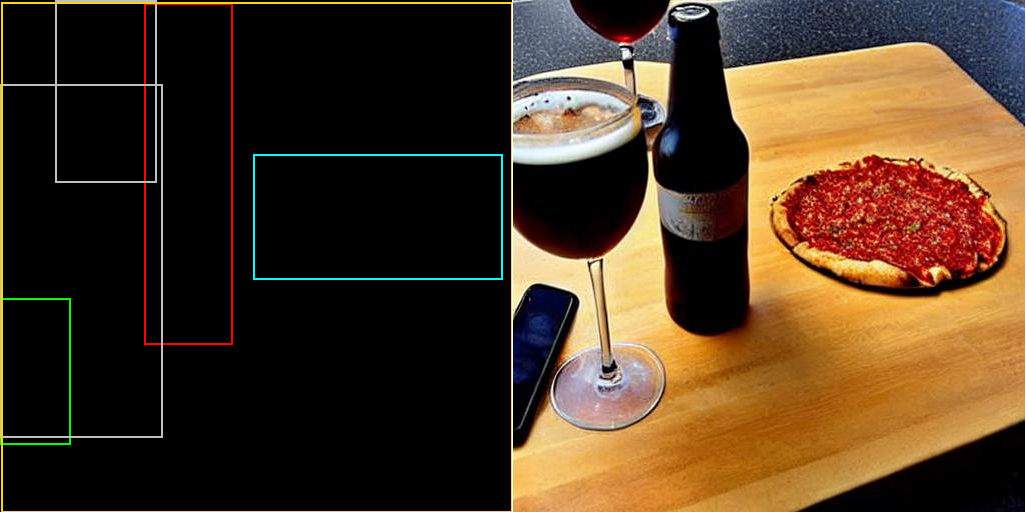}
\subcaption[second caption.]{
\underline{Image Caption:} “Pizza on a cutting board is next to goblets with a beer bottle" \\\underline{Conditional Entities:} \textcolor{myred}{bottle}, \textcolor{mygreen}{cell phone}, \textcolor{mysilver}{wine glass}, \textcolor{mycyan}{pizza}, \textcolor{mygold}{dining table}
}\label{fig:5b}
\end{minipage}%
\hspace{0.001\textwidth}
\begin{minipage}[t]{0.327\textwidth}
  \centering
   \captionsetup{font=small} 
\includegraphics[width=1\textwidth]{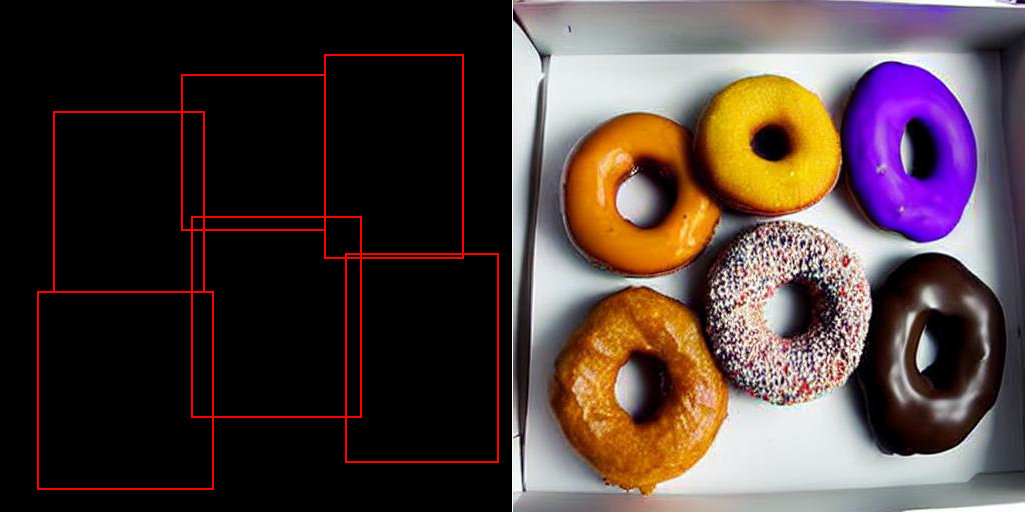}
\subcaption[third caption.]{
\underline{Image Caption:} “A box contains six donuts with varying types of glazes and toppings" \\\underline{Conditional Entities:} \textcolor{myred}{donut}
}\label{fig:5c}
\end{minipage}

\vspace{0.3cm}

\centering
\begin{minipage}[t]{0.327\textwidth}
  \centering
  \captionsetup{font=small} 
\includegraphics[width=1\textwidth]{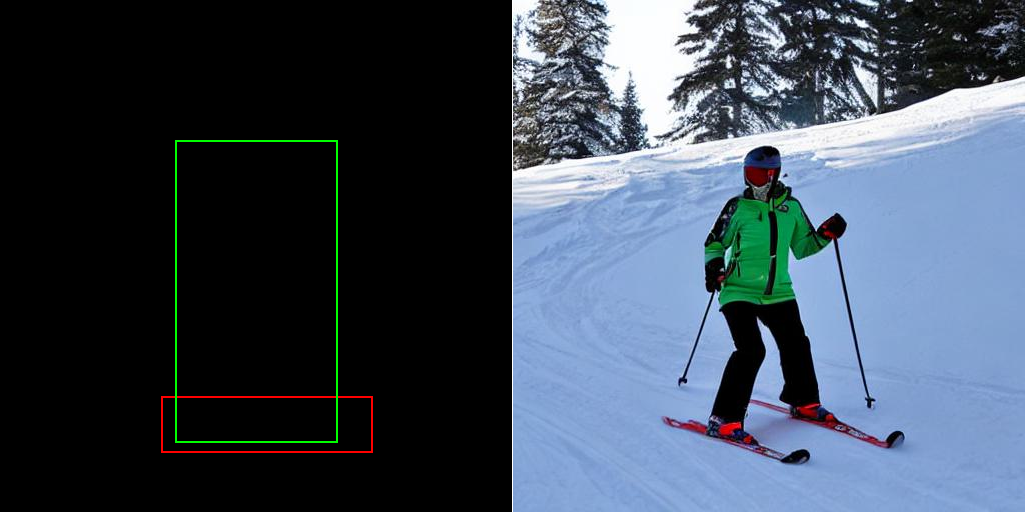}
\subcaption[first caption.]{
\underline{Image Caption:} “A person is trying to ski down a snowy slope" \\\underline{Conditional Entities:} \textcolor{myred}{skis}, \textcolor{mygreen}{person}
}\label{fig:5a}
\end{minipage}%
\hspace{0.001\textwidth}
\begin{minipage}[t]{0.327\textwidth}
  \centering
  \captionsetup{font=small} 
\includegraphics[width=1\textwidth]{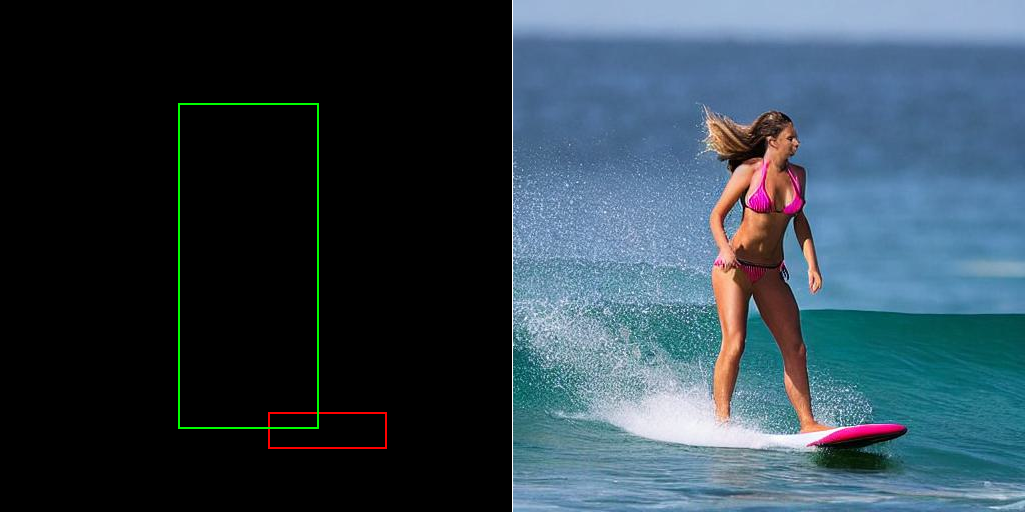}
\subcaption[second caption.]{
\underline{Image Caption:} “A woman in a bikini riding a wave on a surfboard" \\\underline{Conditional Entities:} \textcolor{myred}{surfboard}, \textcolor{mygreen}{person}
}\label{fig:5b}
\end{minipage}%
\hspace{0.001\textwidth}
\begin{minipage}[t]{0.327\textwidth}
  \centering
   \captionsetup{font=small} 
\includegraphics[width=1\textwidth]{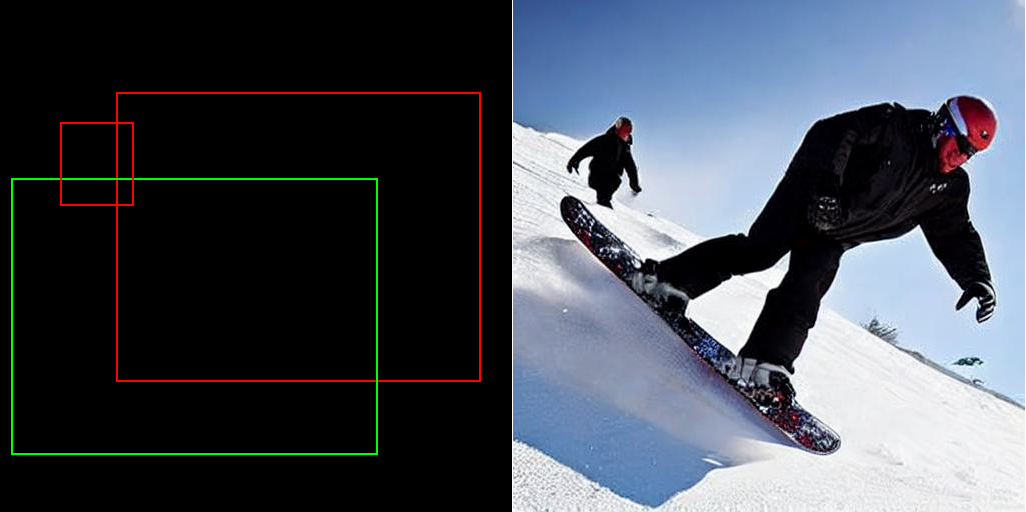}
\subcaption[third caption.]{
\underline{Image Caption:} “A man riding a snowboard on top of a snow covered slope" \\\underline{Conditional Entities:} \textcolor{myred}{person}, \textcolor{mygreen}{snowboard}
}\label{fig:5c}
\end{minipage}

\caption{This figure represents qualitative examples of closed-set configuration. It shows both static and dynamic scenes of different object categories. ObjectDiffusion can synthesize images of dishes as well as visuals of sports activities, such as surfing and skiing. Images are generated using grounding entities and image captions from COCO2017 \cite{lin2014microsoft} validation set annotations.}
\label{fig:closed-set-various}
\end{figure*}

\textbf{Closed-set Setting Evaluation:}
Figures \ref{fig:closed-set-indoor} to \ref{fig:closed-set-various} display generated images under the closed-set vocabulary setting. Entity names and bounding boxes, used as control inputs to synthesize images for the closed-set vocabulary evaluation, are derived from the annotations in the COCO2017 validation set. Each figure showcases a specific general concept, such as people, animals, vehicles, and rooms.

Figure \ref{fig:closed-set-indoor} displays various photo-realistic rooms from a standard home, showcasing the ability of ObjectDiffusion to synthesize high-quality, precise images from complex layouts. For example, the layout of image (c) in the first row of Figure \ref{fig:closed-set-indoor} is an object-intensive visual of living room objects, where the layout contains 19 size-varying bounding boxes labeled with 8 entity classes, such as couch, chair, vase, and TV.

Figure \ref{fig:closed-set-vehicles} shows a collection of land vehicles, railway vehicles, and watercraft. The second row of Figure \ref{fig:closed-set-vehicles} demonstrates the capabilities of our model in adapting to bounding boxes with varying aspect ratios. ObjectDiffusion successfully accommodates motorcycles and persons in each of the three instances with respect to the grounding entity locations and sizes. Additionally, the output images adhere to the image captions, placing the motorcycle on a paved track in the left image (d), on a dirt track in the middle image (e), and on a highway in the right image (f).

\begin{table}[htbp]
    \centering
    \begin{tabular}{l|c}
        \hline 
        \multirow{2}{*}{Model} & \multicolumn{1}{c}{YOLOv8m \cite{redmon2016you}} \\
        & AR (\(\uparrow\)) \\
        \hline upper-bound  & 49.7 \\
        \hline GLIGEN \cite{li2023gligen}  & 30.7 \\
        ObjectDiffusion (ours) & $\mathbf{44.5}$ \\
        \hline
    \end{tabular}
    \caption{This table represents the evaluation of the ability of ObjectDiffusion to correctly render objects specified in the condition layout. ObjectDiffusion demonstrates an outstanding ability to generate the prescribed entities and overcome the object overlooking problem that some text-to-image and layout-to-image models suffer from. The AR \cite{padilla2020survey} scores are calculated using a pretrained YOLOv8m \cite{redmon2016you} object detection model on synthesized images from the COCO2017 \cite{lin2014microsoft} validation set annotations with an image size of 512 $\times$ 512.}
    \label{tab:AR_scores}
\end{table}

In Figure \ref{fig:closed-set-people}, we present synthesized images of people in formal and casual situations. ObjectDiffusion renders photo-realistic portraits with authentic facial features. We observe that in image (c) of Figure \ref{fig:closed-set-people}, the person is wearing sunglasses but is missing the black hat specified in the prompt. We hypothesize that the model may not have encountered enough instances of men in suits with hats on their heads. Additionally, we note that in the same image, the tie is not fully positioned within the bounding box. We also observe that the hands of people in images (b) and (c) are not perfectly rendered. This limitation is inherited from the pretrained SD \cite{rombach2022high} model weights from which we initialized our model.

We exhibit instances of domestic and wild animals in Figure \ref{fig:closed-set-animals}, demonstrating artificial images of dogs, cats, horses, and bears. The high grounding precision of qualitative examples from different animal categories aligns with the high AP scores obtained in the animal categories reported in Table \ref{tab:AP_scores_class}.

The final examples of visuals generated under closed-set vocabulary settings are presented in Figure \ref{fig:closed-set-various}. This figure depicts various images of teddy bears, vases, food, and sporting activities.

ObjectDiffusion can generate instances from a single grounding box or multiple grounding boxes, such as the teddy bear in images (a), (b), and (c) of Figure \ref{fig:closed-set-various}. We can utilize the image prompt as a global condition that describes the overall surroundings and background. For example, images (b) and (c) in Figure \ref{fig:closed-set-various} show a teddy bear laying on a wall and teddy bears sitting on a sofa. In some cases, our model may overlook the colors or materials of the conditional objects, such as the color of the vase in image (f). The grounding ability of ObjectDiffusion is not limited to static visuals. The last row of Figure \ref{fig:closed-set-various} illustrates that our model can render objects in dynamic scenes, such as skiing and surfing, as shown in examples (j), (k), and (l).

\begin{figure*}[htbp]

\centering
\begin{minipage}[t]{0.327\textwidth}
  \centering
  \captionsetup{font=small} 
\includegraphics[width=1\textwidth]{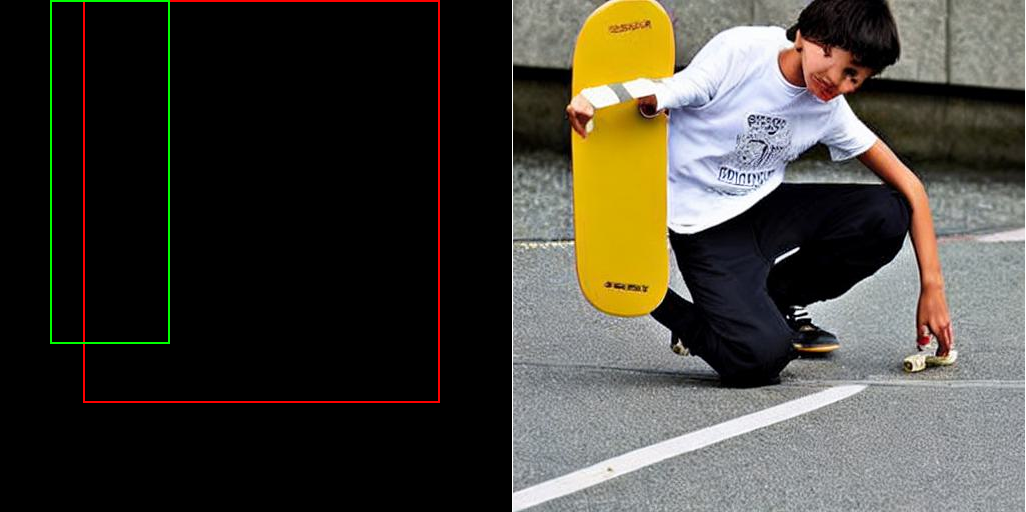}
\subcaption[first caption.]{
 \underline{Image Caption:} “a young man is on his skateboard doing a trick" \\\underline{Conditional Entities:} \textcolor{myred}{person}, \textcolor{mygreen}{skateboard}}\label{fig:5a}
\end{minipage}%
\hspace{0.001\textwidth}
\begin{minipage}[t]{0.327\textwidth}
  \centering
  \captionsetup{font=small} 
\includegraphics[width=1\textwidth]{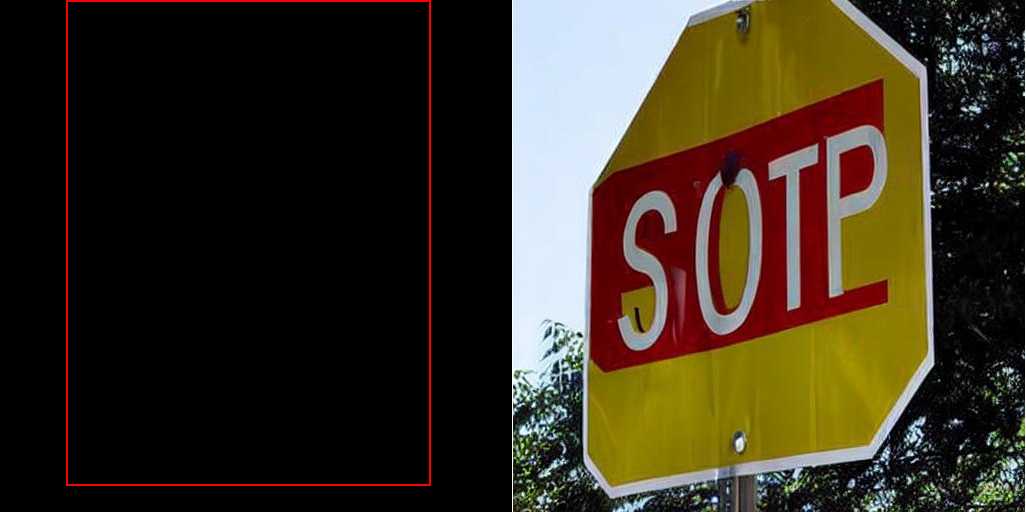}
\subcaption[second caption.]{
 \underline{Image Caption:} “A stop sign is shown among foliage and grass" \\\underline{Conditional Entities:} \textcolor{myred}{stop sign}
}\label{fig:5b}
\end{minipage}%
\hspace{0.001\textwidth}
\begin{minipage}[t]{0.327\textwidth}
  \centering
   \captionsetup{font=small} 
\includegraphics[width=1\textwidth]{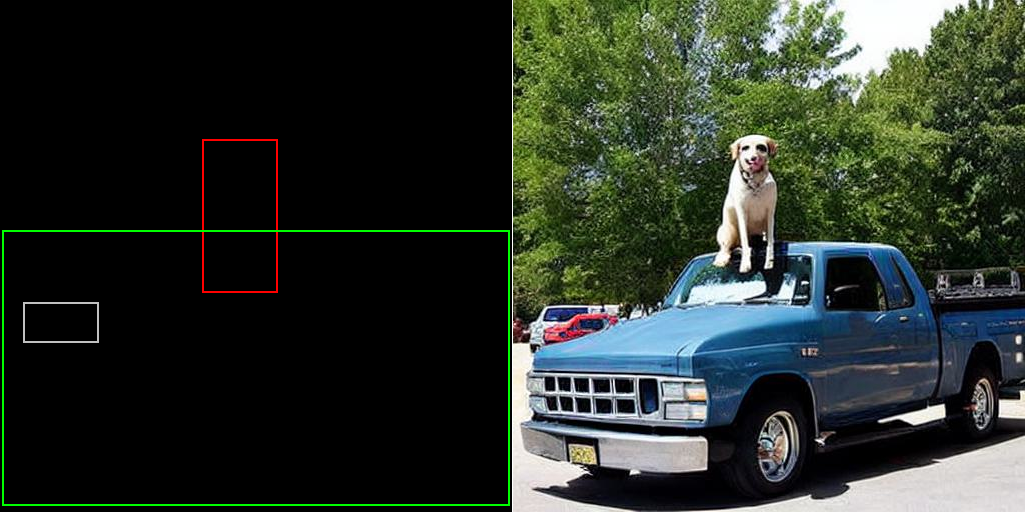}
\subcaption[third caption.]{
 \underline{Image Caption:} “There is a dog on top of a truck outside" \\\underline{Conditional Entities:} \textcolor{myred}{dog}, \textcolor{mygreen}{truck}, \textcolor{mysilver}{car}
 }\label{fig:5c}
\end{minipage}
\caption{This figure showcases limitations of our model in generating faces and hands of people, as shown in image (a). It also highlights the limitations of rendering legible, correct text, as seen in image (b). ObjectDiffusion may synthesize images that precisely adhere to the conditional bounding boxes but have irrational position alignments between objects. For example, in image (c), the dog is absurdly floating above the truck. The images are generated using grounding entities and image captions from the COCO2017 \cite{lin2014microsoft} validation set annotations.}
\label{fig:failure-cases}
\end{figure*}

\begin{figure*}[htbp]

\centering
\begin{minipage}[t]{0.9999\textwidth}
  \centering
  \captionsetup{font=small} 
\includegraphics[width=1\textwidth]{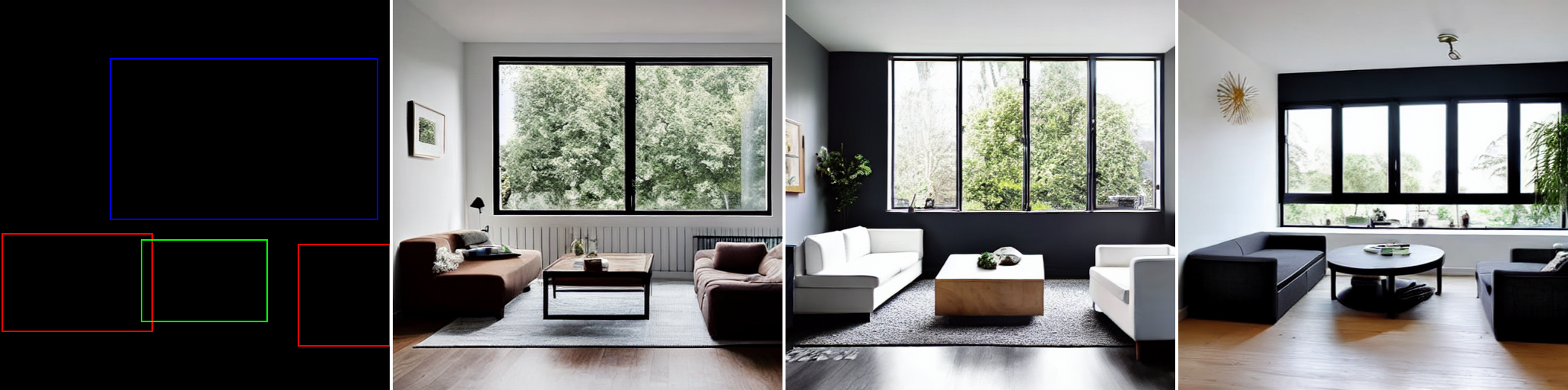}
\subcaption[first caption.]{
\underline{Image Caption:} “bright modern Scandinavian style house with a comfy couch, stylish coffee table, and sunlit windows homevibes relaxation" \\\underline{Conditional Entities:} \textcolor{myred}{a comfy couch}, \textcolor{mygreen}{a stylish coffee table}, \textcolor{myblue}{a sunlit windows}
}\label{fig:6a}
\end{minipage}%

\vspace{0.2cm}
\centering
\begin{minipage}[t]{0.9999\textwidth}
  \centering
  \captionsetup{font=small} 
\includegraphics[width=1\textwidth]{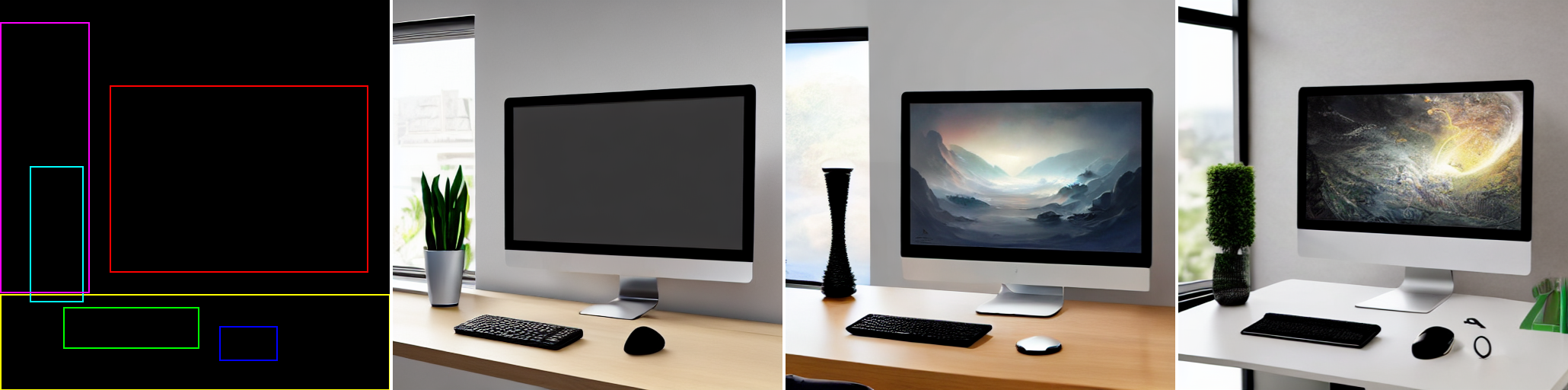}
\subcaption[first caption.]{
\underline{Image Caption:} “a sleek, ultra-thin, high resolution bezel-less monitor, realistic, modern, detailed, vibrant colors, glossy finish, floating design, lowlight, art by peter mohrbacher and donato giancola, digital illustration, trending on Artstation, high-tech, smooth, minimalist workstation background, crisp reflection on screen, soft lighting" \\\underline{Conditional Entities:} \textcolor{myred}{a sleek, ultra-thin, high resolution bezel-less monitor}, \textcolor{mygreen}{a keyboard}, \textcolor{mysilver}{a mouse}, \textcolor{mycyan}{a studying desk}, \textcolor{mygold}{a vase}, \textcolor{mymagenta}{a window}}\label{fig:6a}
\end{minipage}%

\vspace{0.2cm}

\centering
\begin{minipage}[t]{0.9999\textwidth}
  \centering
  \captionsetup{font=small} 
\includegraphics[width=1\textwidth]{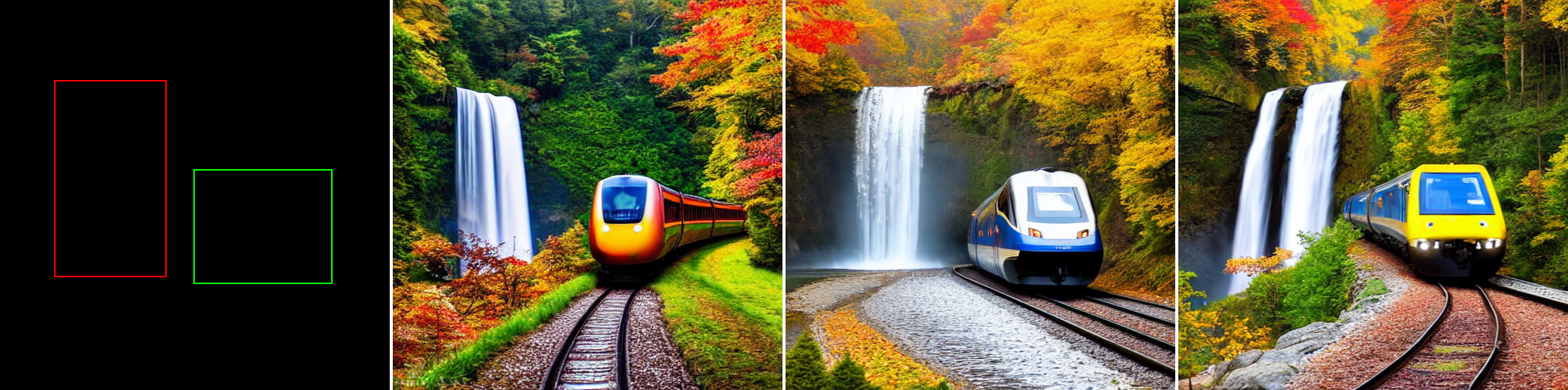}
\subcaption[first caption.]{
\underline{Image Caption:} “a waterfall and a modern high speed train running through the tunnel in a beautiful forest with fall foliage" \\\underline{Conditional Entities:} \textcolor{myred}{a waterfall}, \textcolor{mygreen}{a modern high speed train}}\label{fig:6a}
\end{minipage}%

\vspace{0.2cm}

\centering
\begin{minipage}[t]{0.9999\textwidth}
  \centering
  \captionsetup{font=small} 
\includegraphics[width=1\textwidth]{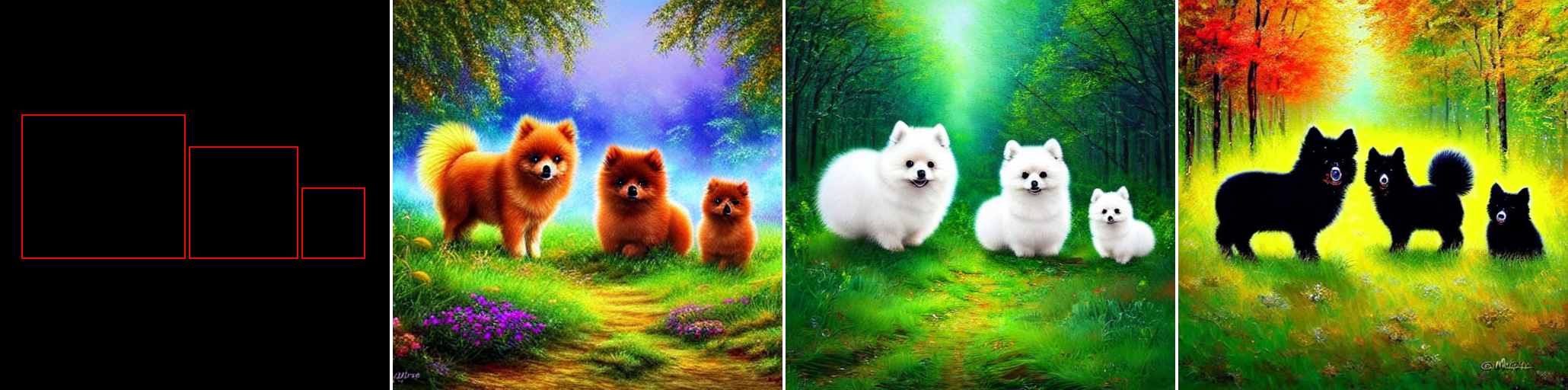}
\subcaption[first caption.]{
\underline{Image Caption:} “a beautiful digital painting depecting fluffy Pomeranian in a green forest, impressionism painting style, impressive, masterpeace"   \\\underline{Conditional Entities:} \textcolor{myred}{a fluffy Pomeranian}}\label{fig:6a}
\end{minipage}%

\caption{This figure represents realistic and artistic qualitative examples of images generated in an open-set configuration with manual annotations. The images demonstrate the high quality and precision of ObjectDiffusion.}
\label{fig:open-set-various}
\end{figure*}

\textbf{Open-set Setting Evaluation:}
We evaluate the performance of ObjectDiffusion in grounding the generation on entity names that are not present in the fine-tuning dataset. To achieve this, we manually craft the captions, bounding boxes, and entity descriptions using entity names outside the 80 COCO classes. In Figure \ref{fig:open-set-various}, we provide three examples for each layout.

\begin{table}[htbp]
    \centering
    \begin{tabular}{l|c}
        \hline 
        \multirow{2}{*}{Model} & \multirow{2}{*}{FID (\(\downarrow\))} \\
        & \\
        \hline LostGAN-V2 \cite{sun2021learning} & 42.55 \\
        OCGAN \cite{sylvain2021object} & 41.65 \\
        HCSS \cite{jahn2021high} & 33.68 \\
        LAMA \cite{li2021image} & 31.12 \\
        TwFA \cite{yang2022modeling} & 22.15 \\
        GLIGEN \cite{li2023gligen} & 21.04 \\
        \hline \multicolumn{1}{c}{ pretrained on (Object365, GoldG, SBU, CC3M) } \\
        GLIGEN (zero-shot) \cite{li2023gligen} & 27.03 \\
        GLIGEN (fine-tuned) \cite{li2023gligen} & 21.58 \\
        ObjectDiffusion (ours)& $\mathbf{19.80}$ \\
        \hline
        \end{tabular}
        \caption{This table lists the quality evaluation results of generated images from the COCO2017 \cite{lin2014microsoft} validation set. ObjectDiffusion achieves the highest quality, setting a new benchmark in conditional image generation. We use the FID \cite{heusel2017gans} score implementation from \cite{Seitzer2020FID} and resize both the real and generated images to 299 $\times$ 299 before calculating the FID score. Other FID scores are taken from \cite{li2023gligen}.}
    \label{tab:FID_scores}
\end{table}

The first row (a) demonstrates image examples of a modern house in Scandinavian style. We can use the category names from the 80 COCO categories but with modifier attributes preceding the name, such as comfy couch, stylish coffee table, or we can use completely new concepts, such as sunlit windows. We observe a noticeable improvement in the quality of images in the open-set vocabulary setting compared to their counterparts in the closed-set vocabulary setting. This enhancement may be explained by the flexibility that the open-ended text provides.

Second row (b) depicts visuals of fancy desk setups. ObjectDiffusion can understand long and complex captions and entity texts, such as a sleek, ultra-thin, high-resolution bezel-less monitor. Also, our model was creative in rendering the three vases, each with a distinct shape.

Third row (c) contains stunning images of a modern high-speed train passing beside a waterfall. Even though the waterfall is an uncommon and has never been seen as an instance name in the COCO2017 dataset, our model is able to interpret and generate it. However, we observe that all of our output images miss the tunnel. This may be because it was not in the grounding entity input. The open-set qualitative visuals indicate that ObjectDiffusion is able to retain the knowledge of the pretrained SD \cite{rombach2022high} and GLIGEN \cite{li2023gligen} models while successfully demonstrating conditioning capabilities on the fine-tuning grounding task.

The last row (d) in Figure \ref{fig:open-set-various} emphasizes the powerful regional grounding skills of our model. The layout is constructed by stacking three bounding boxes of varying sizes, all belonging to the same object name, a fluffy Pomeranian. As seen in the synthesized images, ObjectDiffusion succeeds in generating the correct entity with the correct size in the correct location. Furthermore, our model exhibits a high level of diversity, rendering the Pomeranians in three different colors: brown, white, and black, and with three different background illumination contrasts.

Generated images from the open-set vocabulary settings demonstrate the capabilities ObjectDiffusion of synthesizing realistic and artistic visuals, both of which are spectacular, diverse, and high-quality.

\subsection{Limitations and Failures}

ObjectDiffusion is able to synthesize high-quality scenes; however, it sometimes struggles with capturing the intricate fine-grained details of faces, hands, and feet \cite{rosenberg2024limitations}. These are highly detailed, complex structures that exhibit a high degree of variability. Image (a) in Figure \ref{fig:failure-cases} is an example of a young man with a distorted face and hands.

Similar to the limitation in face synthesis, ObjectDiffusion exhibits an imperfection in rendering images that contain text. We observe that our model repeatedly fails to render images with embedded text. As seen in image (b) in Figure \ref{fig:failure-cases}, our model fails to overlay a short, single-word text on the stop sign, even though our fine-tuning dataset contains an abundance of stop sign images.

ObjectDiffusion may fail to account for the spatial relationships and interactions between objects in the scene. This leads to unrealistic object placements. For instance, in image (c) of Figure \ref{fig:failure-cases}, while the dog and the vehicle are correctly positioned within their designated bounding boxes, the legs of the dog do not make contact with the truck, creating an unnatural scene where the dog seems to float, disconnected from its surroundings.

\section{Discussion and Conclusion}
In this work, we draw inspiration from two influential models, ControlNet and GLIGEN, to propose a new model, ObjectDiffusion. Our model learns to ground T2I models on semantic and spatial controls. ObjectDiffusion outperforms current state-of-the-art models trained on open-source datasets in AP$_{\text{50}}$ and AR metrics. Additionally, ObjectDiffusion scores higher than the top-performing layout-to-image baselines in AP and AP$_{\text{75}}$ metrics. Our model sets a new record in the FID quality metric, surpassing the highest scores previously achieved by both layout-to-image models and diffusion-based models conditioned on COCO2017 object detection annotations. Qualitative evaluation highlights the capability of ObjectDiffusion to generate diverse, high-quality, high-fidelity images that consistently align with the provided conditional inputs.

While ObjectDiffusion has demonstrated remarkable performance, several potential improvements could further enhance the capabilities of our model. First, training on a large-scale dataset containing precise and detailed entity names could notably increase the AP score. Second, training our model for more iterations may lead to a noticeable improvement in overall performance. Additionally, training on powerful GPUs would allow for a larger batch size and enable the use of full precision. Scaling up the training has the potential to accelerate model convergence and further improve generation precision and quality.

\section*{Acknowledgment}
We are deeply grateful to Thomas J. Crabtree for his valuable assistance with language editing and spelling corrections.

\clearpage

\clearpage
\small
\bibliographystyle{ieeenat_fullname}
\bibliography{main}

\cleardoublepage


\end{document}